\documentclass[letterpaper]{article} 
\usepackage{aaai2026}
\usepackage{times}  
\usepackage{helvet}  
\usepackage{courier}  
\usepackage[hyphens]{url}  
\usepackage{graphicx} 
\usepackage{natbib}  
\usepackage{caption} 
\usepackage{algorithm}
\usepackage{algorithmic}

\usepackage{tabularx}
\usepackage{subcaption}
\usepackage{makecell}
\usepackage{placeins}
\usepackage{utfsym}
\usepackage{afterpage}
\usepackage{url}
\usepackage{colortbl}
\usepackage[table]{xcolor} 
\usepackage{booktabs}
\usepackage{multirow}
\usepackage{amsmath}

\usepackage[most]{tcolorbox} 
\usepackage{subcaption}
\usepackage{makecell}
\usepackage{placeins}
\usepackage{utfsym}
\usepackage{afterpage}
\usepackage{url}
\usepackage{colortbl}
\usepackage[table]{xcolor} 
\usepackage{booktabs}
\usepackage{multirow}
\usepackage{newfloat}
\usepackage{listings}
\DeclareCaptionStyle{ruled}{labelfont=normalfont,labelsep=colon,strut=off} 
\floatstyle{ruled}
\newfloat{listing}{tb}{lst}{}
\floatname{listing}{Listing}
\title{CultureScope: A Dimensional Lens for Probing Cultural Understanding in LLMs}
\author{
    Jinghao Zhang\textsuperscript{\rm 1},
    Sihang Jiang\textsuperscript{\rm 1, *},
    Shiwei Guo\textsuperscript{\rm 2},
    Shisong Chen\textsuperscript{\rm 3},
    Yanghua Xiao\textsuperscript{\rm 1, *},
    Hongwei Feng\textsuperscript{\rm 1},
    Jiaqing Liang\textsuperscript{\rm 1},
    Minggui HE\textsuperscript{\rm 4},
    Shimin Tao\textsuperscript{\rm 4},
    Hongxia Ma\textsuperscript{\rm 4} %
    \thanks{Corresponding authors.}
}

\affiliations{
    \textsuperscript{\rm 1}Shanghai Key Laboratory of Data Science, College of Computer Science and Artificial Intelligence, Fudan University\\

    \textsuperscript{\rm 2}School of Foreign Languages and Literatures, Fudan University\\

    \textsuperscript{\rm 3}Shanghai Institute of Artificial Intelligence for Education, East China Normal University\\

    \textsuperscript{\rm 4}Huawei, China\\

    zhangjinghao24@m.fudan.edu.cn, \{jiangsihang, shawyh\}@fudan.edu.cn
}

\usepackage{bibentry}

\begin{document}

\maketitle

\begin{abstract}
As large language models (LLMs) are increasingly deployed in diverse cultural environments, evaluating their cultural understanding capability has become essential for ensuring trustworthy and culturally aligned applications. 
However, most existing benchmarks lack comprehensiveness and are challenging to scale and adapt across different cultural contexts, because their frameworks often lack guidance from well-established cultural theories 
and tend to rely on expert-driven manual annotations.
To address these issues, we propose \textit{CultureScope}, the most comprehensive evaluation framework to date for assessing cultural understanding in LLMs.
Inspired by the cultural iceberg theory, we design a novel dimensional schema for cultural knowledge classification, comprising 3 layers and 140 dimensions, which guides the automated construction of culture-specific knowledge bases and corresponding evaluation datasets for any given languages and cultures.
Experimental results demonstrate that our method can effectively evaluate cultural understanding. They also reveal that existing large language models lack comprehensive cultural competence, and merely incorporating multilingual data does not necessarily enhance cultural understanding. All code and data files are available at \url{https://github.com/HoganZinger/Culture}

\end{abstract}


\section{Introduction}
\begin{figure}[t]
\centering
  \includegraphics[scale=0.35]{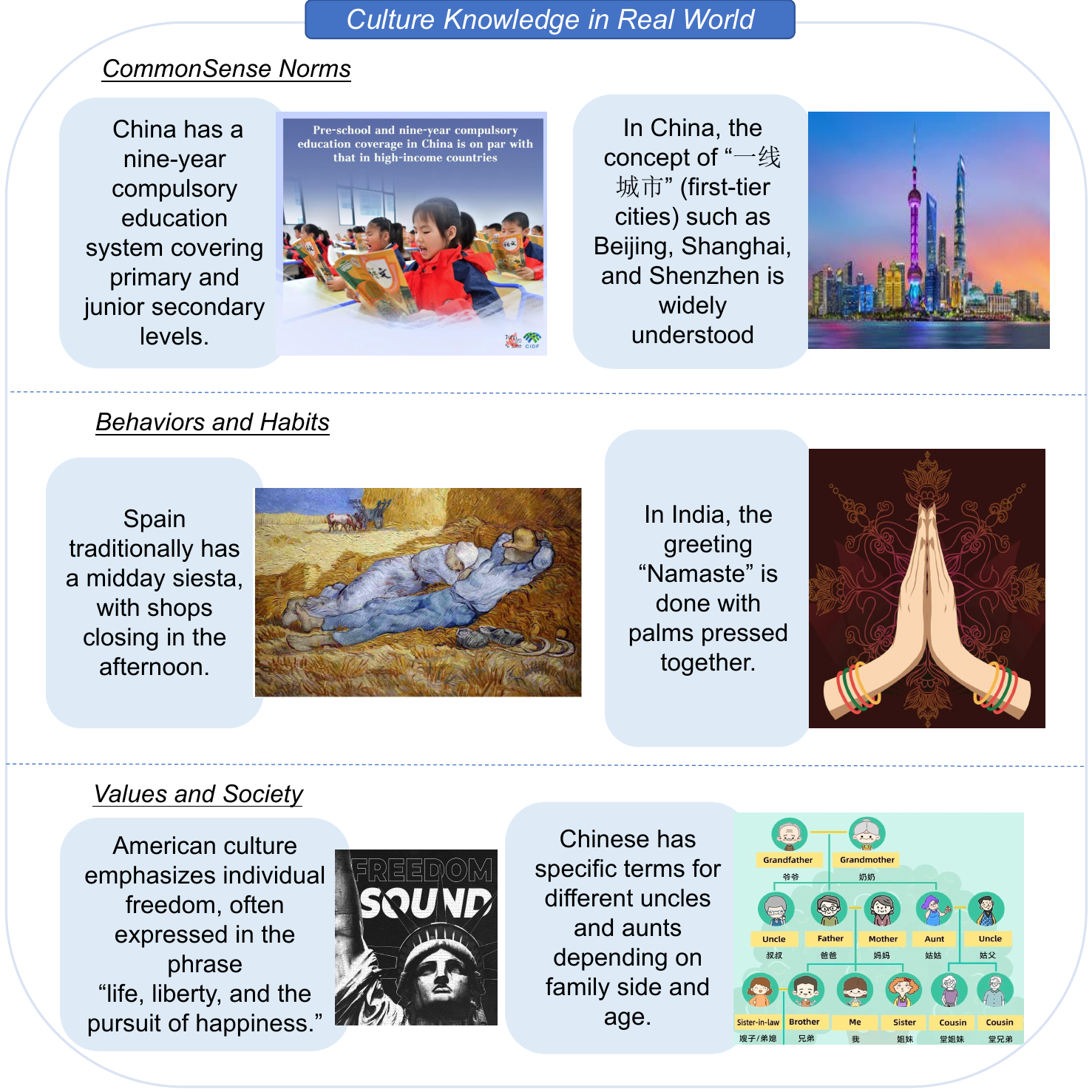}
  \caption{The classification of real-world cultural knowledge is inherently complex and diverse, encompassing a wide range of aspects such as institutions, geography, behavioral habits, etiquette, values, and social structures. Existing evaluation datasets often fail to comprehensively and accurately cover these types of knowledge.
}
  \label{fig: figure1}
\end{figure}

Large language models (LLMs) have demonstrated impressive capabilities across a wide spectrum of natural language processing tasks, and are now widely adopted in applications ranging from virtual assistants to educational tools.\cite{Mohan2024An,raiaan2024review,raza2025industrial} Despite these achievements, a persistent challenge remains: Current large language models still exhibit significant shortcomings in cultural understanding, often leading to cultural misalignment when applied in real-world contexts\cite{Benkler2023Assessing,AlKhamissi2024Investigating,Shen2024Understanding}. For example, if a healthcare chatbot powered by an LLM recommended placing an elderly parent in a nursing facility to a Chinese user, it fails to account for the cultural significance of filial piety (Xiaoshun) in traditional Chinese society. While the suggestion aligns with Western notions of efficiency, it conflicts with deeply rooted Chinese values, leading to user dissatisfaction and negative public response. 
Therefore, how to comprehensively and accurately evaluate LLMs' cultural understanding abilities in specific cultural contexts has become an urgent issue that must be addressed for their broader adoption.

\begin{figure*}[t]
\centering
  \includegraphics[scale=0.3]{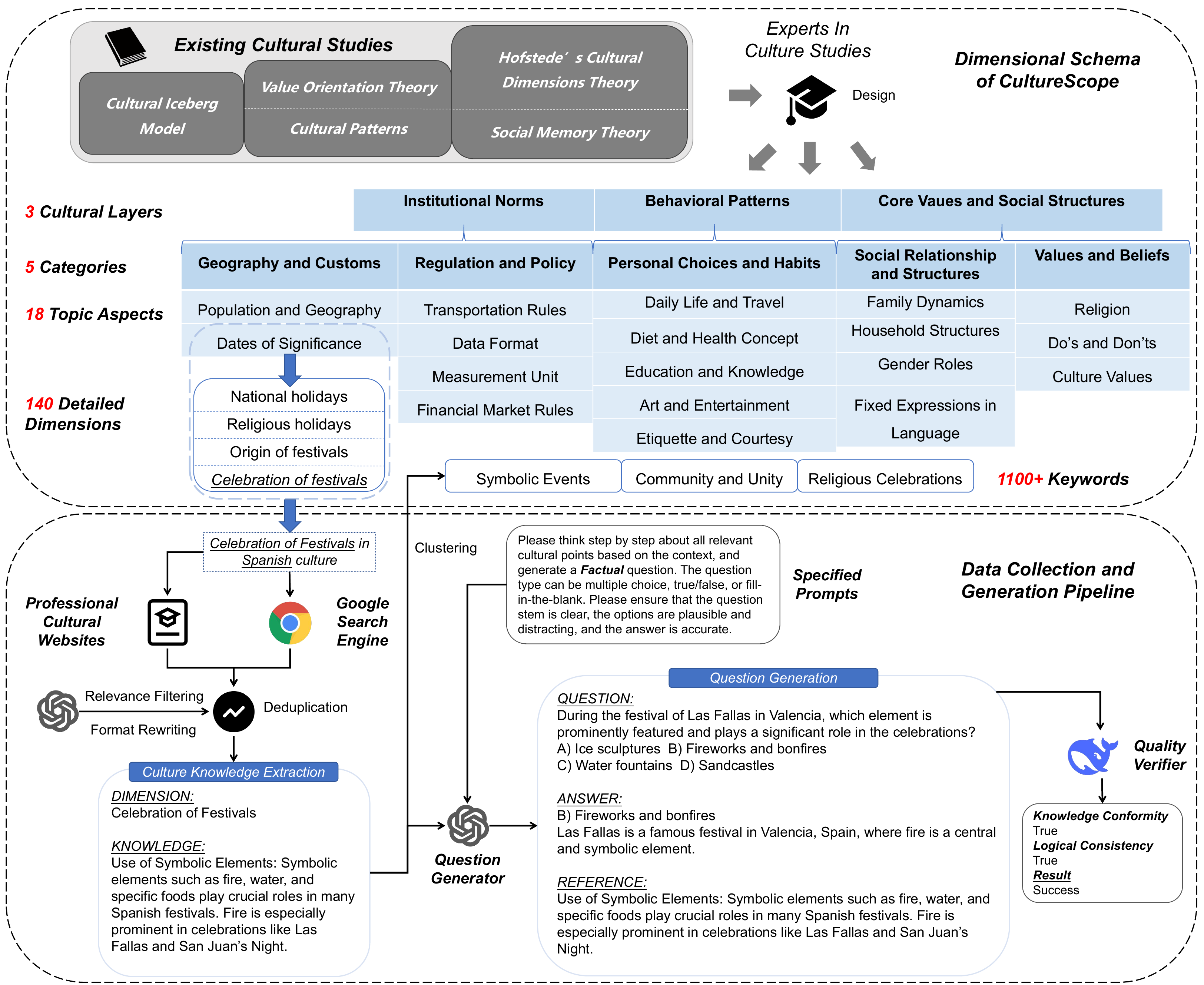}
  \caption{The overall framework of CultureScope. The framework can be applied to the evaluation of cultural understanding ability in any target culture. We invite experts in culture studies to to develop the most comprehensive knowledge classification schema to date for evaluating the cultural understanding of large language models, based on the most influential existing cross-cultural and mono-cultural theories. This framework 
  enables more fine-grained analysis and interpretation of LLMs’ performances on specific cultural dimensions.}
  \label{fig:generation}
\end{figure*}

Cultural understanding refers to a model’s ability to comprehend culturally grounded definitions and concepts, identify cultural stereotypes or biases, and appropriately apply cultural knowledge in real-world or context-sensitive scenarios\cite{Earley2002Redefining}.
As is shown in Figure~\ref{fig: figure1}, culture is a complex and holistic concept\cite{tylor1871primitive}, and cultural knowledge in the real world scenarios encompasses multiple dimensions. 
Despite growing efforts in the evaluation and enhancement of cultural understanding, existing benchmarks still suffer from significant limitations. 


\textbf{First}, existing evaluations tend to lack comprehensiveness, which limits their practical utility, and the assessed capability scores may be misaligned with real-world risks and alignment requirements. Previous studies either directly rewrite existing surveys\cite{jiang2024can,zhao2024worldvaluesbench,li2024culturellm}, or generate and manually annotate data by selecting a small number of keywords\cite{chiu2024culturalbench,lee2024kornat,myung2024blend}. 
Their classification frameworks for cultural knowledge dimensions lack rigorous theoretical guidance and tend to rely on intuitive and empirical experience. 
\textbf{Second}, existing evaluation benchmarks are difficult to scale and adapt to other cultural and linguistic contexts, as they heavily rely on manual annotation\cite{myung2024blend,palta2023fork,mousi2024aradice}. The high cost, labor intensity, and cultural specificity of manual labeling limit their scalability and hinder their generalizability across diverse cultural settings. \textbf{Moreover}, many existing evaluations mainly test cultural knowledge, but a comprehensive assessment should also examine whether models can recognize cultural bias and apply cultural knowledge in context, closer to real-world scenarios. 

\newcolumntype{Y}{>{\centering\arraybackslash}X}
\begin{table*}[htbp]
  \centering
  \renewcommand{\arraystretch}{1.3} 
  \begin{tabularx}{\textwidth}{|c|c|c|c|c|c|Y|}
    \hline
    \textbf{Benchmark} 
    & \begin{tabular}[c]{@{}c@{}}\textbf{Theoretical}\\ \textbf{Guidance}\end{tabular} 
    & \begin{tabular}[c]{@{}c@{}}\textbf{Commonsense}\\ \textbf{Norms}\end{tabular} 
    & \begin{tabular}[c]{@{}c@{}}\textbf{Behaviors}\\ \textbf{\& Habits}\end{tabular} 
    & \begin{tabular}[c]{@{}c@{}}\textbf{Values}\\ \textbf{\& Society}\end{tabular} 
    & \begin{tabular}[c]{@{}c@{}}\textbf{Transferability}\\ \textbf{to other cultures} \end{tabular} 
    & \begin{tabular}[c]{@{}c@{}}\textbf{Number of}\\ \textbf{dimensions}\end{tabular} \\
    \hline
    BLEnD(\citeyear{myung2024blend})       
    & \usym{2717}
    & \usym{2717}  & \usym{2713} & \usym{2717}  & \usym{2717} & 6 \\
    \hline
    CultureLLM(\citeyear{li2024culturellm})  & \usym{2717} & \usym{2717} & \usym{2717} & \usym{2713}  & \usym{2717} & 7 \\
    \hline
    WorldValuesBench(\citeyear{zhao2024worldvaluesbench}) & \usym{2717} & \usym{2717} & \usym{2717} & \usym{2713} & \usym{2717} & 12 \\
    \hline
    KorNAT (\citeyear{lee2024kornat})     & \usym{2717} & \usym{2717} & \usym{2713}  & \usym{2713} & \usym{2717} & /  \\
    \hline
    ARADICE  (\citeyear{mousi2024aradice})   & \usym{2717} & \usym{2713} & \usym{2713} & \usym{2713} & \usym{2717}  & 30 \\
    \hline
    
    Ours(CultureScope) & \usym{2713} & \usym{2713} & \usym{2713} & \usym{2713}  & \usym{2713} & \textbf{140} \\
    \hline
  \end{tabularx}
  \caption{
  Existing benchmarks lack guidance from well-established cultural theories, with dimension systems largely designed based on empirical heuristics. As a result, they provide limited coverage of cultural knowledge, hindering comprehensive evaluation of culture understanding.
  Moreover, the reliance on manual annotation in many existing methods leads to limited automation and poses significant challenges in extending them to diverse cultural contexts.
  }
  \label{tab:comparisons}
\end{table*}
In this paper, we propose CultureScope, an automated, scalable, and currently the most comprehensive evaluation framework for LLMs’ cultural understanding capabilities. 
To address the issues above, we argue that the evaluation framework for cultural understanding 
should meet four key criteria: \textit{comprehensiveness, operability, interpretability, and scalability}. As is shown in Figure~\ref{fig:generation}, to ensure that our framework provides \textbf{comprehensive} coverage of cultural knowledge, we first introduce a dimension-based schema for classifying cultural knowledge. This schema is grounded in existing cultural studies and represents the most comprehensive classification system to date, enabling more systematic, comparable, and fine-grained assessments of large language models’ abilities across specific cultural dimensions. 
Using the sub-dimensions defined in this schema as a foundation, we then \textbf{automatically} extract cultural knowledge instances and construct cultural understanding evaluation datasets for Chinese and Spanish cultures. Furthermore, we \textbf{expand} the schema by clustering the extracted cultural knowledge. 

We conduct extensive evaluations of existing LLMs using our constructed evaluation datasets. 
Our findings indicate that cultural capability fundamentally involves the mastery and application of cultural knowledge. Relying exclusively on linguistic knowledge is insufficient for improving a model’s cultural alignment, and deeper reasoning does not inherently lead to better cultural competence. Furthermore, existing models perform inconsistently across different cultural dimensions and languages, which provides clear guidance for targeted enhancements of the models.



Our contributions are as follows:
\begin{itemize}
\item Building on previous monocultural and cross-cultural research, we propose the most comprehensive framework to date for assessing LLMs’ capabilities in specific cultural contexts. This framework captures both surface-level practices and deep-level assumptions in a certain culture, covering a wide range of cultural knowledge, including cultural facts, behavioral norms, values, and beliefs, enabling automatic data generation and can be flexibly extended to any given culture. 

\item We extract a substantial volume of cultural knowledge embedded in specific cultural contexts from various data sources, which can facilitate subsequent evaluation and training of LLMs. 
These cultural knowledge possess high quality, and the accuracy of knowledge instances retrieved and generated in different languages remain comparable.

\item We construct several comprehensive evaluation datasets and conduct a series of extensive experiments on commonly used LLMs to evaluate their cultural understanding abilities. 
The performances of these LLMs vary significantly across different cultural dimensions and languages. Enhancing models solely through multilingual data and deep reasoning is insufficient to achieve stable improvements in cultural understanding. 
\end{itemize}


\section{Related Work}

\subsection{Culture Understanding Benchmarks}
In recent years, the evaluation and improvement of LLMs’ cultural understanding capabilities have received significant attention within the research community and industry. These capability assessments can broadly be divided into two categories.
(1) Question answering on factual knowledge within specific cultural contexts\cite{onohara2024jmmmu,singh2024global,nayak2024benchmarking}, and (2) Evaluation of cultural values and moral reasoning\cite{hadar2024assessing,karinshak2024llmglobe,aksoy2025whose,ji2025moralbench}. Common data construction methods include manual annotation and LLM generation.
However, current research on the evaluation of LLMs' cultural understanding capabilities still faces notable limitations. Table~\ref{tab:comparisons} provides a detailed comparison between our evaluation benchmark and previous works.
On one hand, most existing evaluation benchmarks are not grounded in robust theoretical frameworks, and their dimensional structures are typically based on \textbf{intuitive, experience-driven approaches}. As a result, their evaluation frameworks fail to cover a broad spectrum of cultural knowledge and cannot fully reveal the limitations and areas for improvement of LLMs in culturally rich scenarios. 
On the other hand, existing methods often rely on \textbf{expert manual annotation}, which incurs high data generation costs and makes it difficult to scale to other cultures. Some datasets even involve hundreds of annotators, which is evidently costly and difficult to replicate.

\subsection{Multilingual Large Language Models}
Multilingual Large Language Models (LLMs)\cite{lai2023chatgpt} can be broadly categorized into two major paradigms. The first paradigm consists of general-purpose multilingual models trained from scratch on massive multilingual corpora\cite{xu2025survey,wang2024multilingual,wei2023polylm}. These models emphasize learning multilingual representations that cover dozenof languages. 
The second paradigm focuses on continued training or fine-tuning of existing general-purpose models to enhance their multilingual capabilities\cite{kadiyala2025improving,toraman2024llamaturk,thillainathan2025beyond}. 
However, whether multilingual training data alone can enable models to fully acquire corresponding cultural understanding remains an open question\cite{Wang2023SeaEval,2024Multilingualism}. Cultural knowledge involves complex social norms and contextual nuances that may not be fully captured by multilingual text corpora\cite{Morris2015Normology}, and further research is needed to evaluate and improve the cultural competence of these models.

\begin{figure*}[htbp]
  \centering
  \begin{subfigure}{0.49\textwidth}
    \includegraphics[height=0.366\textheight, width=\linewidth]{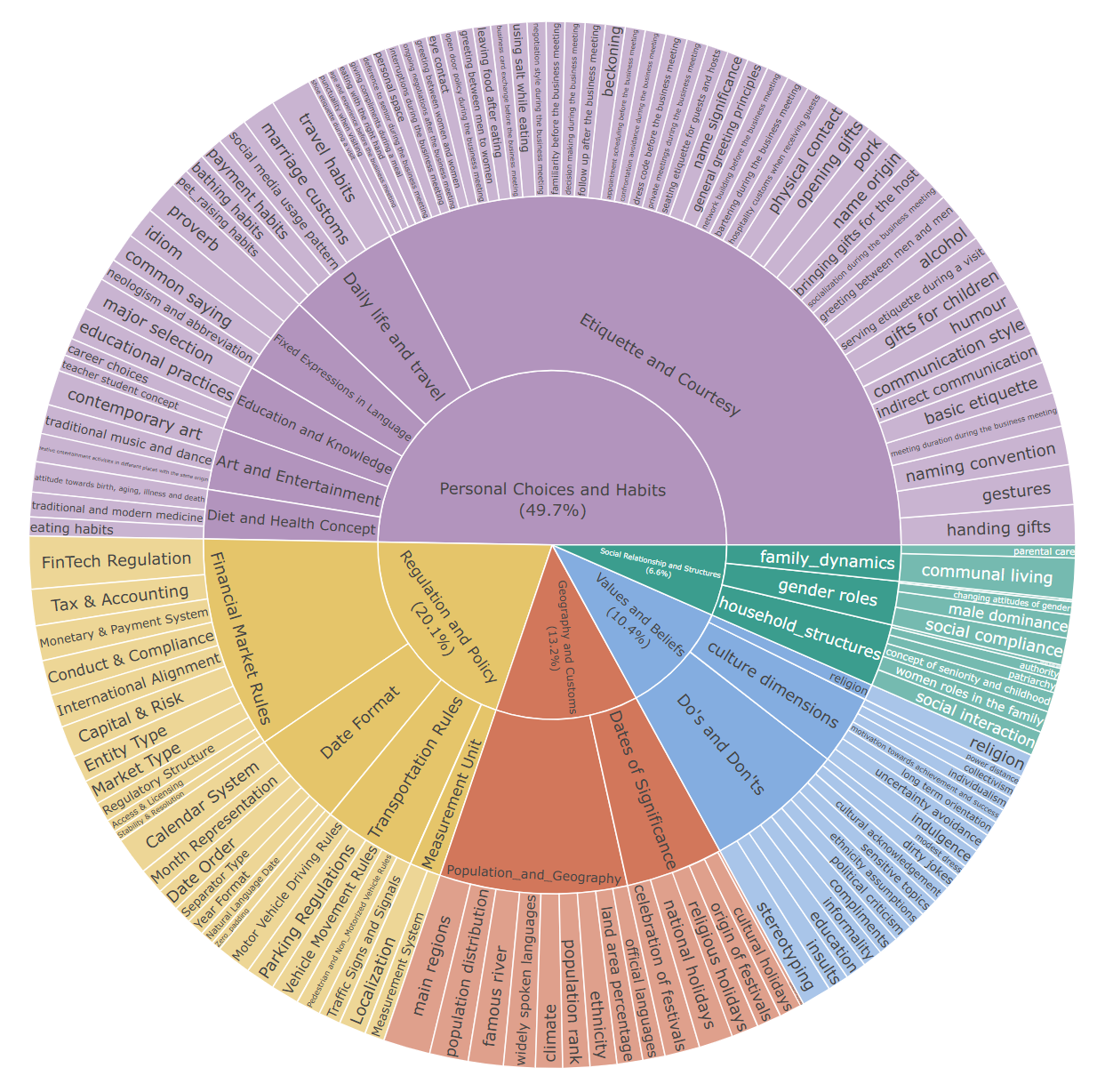}
    \caption{Spanish Cultural Knowledge}
    \label{fig:sub1}
  \end{subfigure}
  \hfill
  \begin{subfigure}{0.49\textwidth}
    \includegraphics[height=0.366\textheight, width=\linewidth]{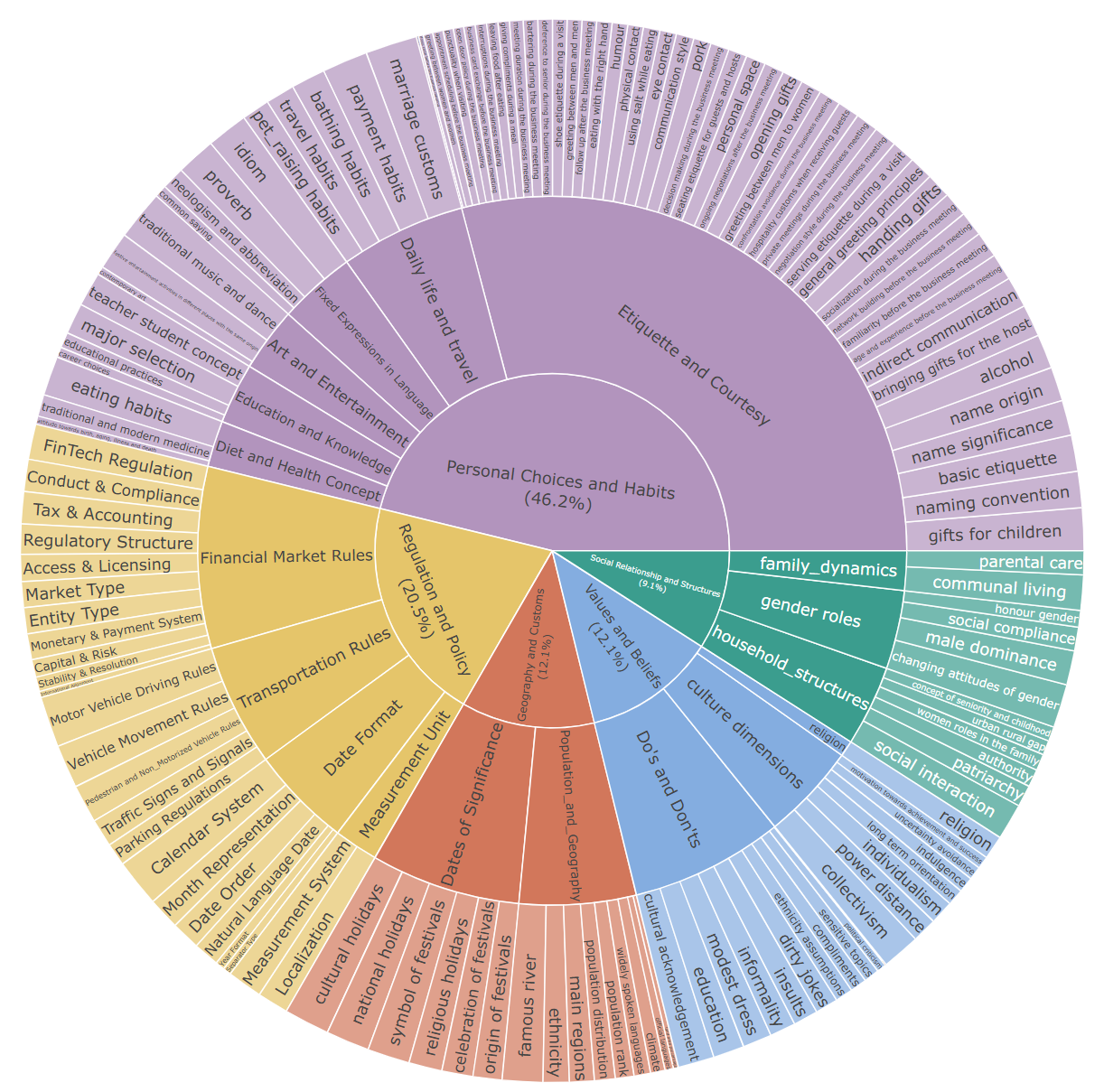}
    \caption{Chinese Cultural Knowledge}
    \label{fig:sub2}
    
  \end{subfigure}

  \caption{Distribution of Retrieved Cultural Knowledge in Spanish and Chinese. These cultural knowledge instances are collected from multiple sources under the guidance of the dimensional schema, and their distribution is consistent with the dimensions defined in the schema.}
  \label{fig:databaseDistribution}
\end{figure*}

\section{CultureScope Framework}
We propose the most comprehensive, automated, and extensive evaluation framework to date for evaluating LLMs' cultural understanding capabilities, which can be flexibly extended to any given cultures. This framework comprises \textbf{3 layers, 5 categories, 18 topic aspects and 140 dimensions}.

\subsection{Dimensional Schema Design}
\paragraph{Classification Principles}
To comprehensively and scientifically evaluate the cultural understanding capability of LLMs, the dimensional framework for classifying cultural knowledge should satisfy four core principles.
\textbf{1) Comprehensiveness}: The dimensions should cover cultural knowledge as fully as possible.
\textbf{2) Operationalizability}: The framework should effectively guide data collection and the practical construction of knowledge bases.
\textbf{3) Interpretability}: It should enable clear attribution analysis of model behaviors, revealing preferences and performance variations across dimensions.
\textbf{4) Extensibility}: The system should be designed for high scalability through data-driven expansion, supporting increasingly fine-grained evaluation in the future.

\paragraph{Dimensional Schema}

Drawing inspiration from the cultural iceberg theory \cite{hall1976beyond,Baker2009The}, we construct a hierarchical classification of cultural knowledge that captures both surface-level practices and deep-level assumptions, dividing it into three layers: Institutional Norms, Behavioral Patterns, and Core Values and Social Structures.
The three layers respectively cover cultural knowledge of common norms (e.g., geography and festivals), behaviors and habits, and underlying values and social structures. As is shown in Figure~\ref{fig:generation}, the schema is divided into four levels: \textit{Culture Layer, Category, Topic Aspect, and Detailed Dimension}, which provides systematic dimensions against which LLMs' cultural understanding can be rigorously assessed. 
The detailed content of the proposed dimensional schema is shown in Appendix.




\begin{table}[]
    \centering
    \renewcommand{\arraystretch}{1.3} 
    \begin{tabular}{ccccc}
    \hline
        \multirow{2}{*}{\textbf{Size}} & \multicolumn{2}{c}{\textbf{Knowledge Instances}} & \multicolumn{2}{c}{\textbf{Evaluation Dataset}}\\
        \cmidrule(lr){2-3}
        \cmidrule(lr){4-5}
        & \textbf{En} & \textbf{Local} & \textbf{En} & \textbf{Local} \\
        \hline
        Spanish & 2767 & 2483 & 11.09k & 10.68k \\
        \hline
        Chinese & 3177 & 3535 & 11.2k& 11.2k \\
    \hline
    \end{tabular}
    \caption{Statistics of CultureScope. We report the number of knowledge instances in each language-specific contexts. The number of data samples is the same across different types of question content(Factual, Conceptual, Misleading, Multi-hop) across languages. 
}
    \label{tab:statistics}
\end{table}

\subsection{Cultural Knowledge Extraction}


Some studies have found that even datasets widely regarded as reliable, such as Wikipedia, still exhibit cultural biases\cite{naous2023having}. Therefore, in our data construction process, we incorporate complementary data sources to ensure diversity and to mitigate the risk of bias from relying on a single source. 
The data construction process of CultureScope takes into account two types of data sources. The first is manually selected professional cultural websites\footnote{For example, \url{https://culturalatlas.sbs.com.au/}}. These sites offer highly reliable content, but the amount of data, the coverage across dimensions is limited, and data extraction often requires specialized processing. The other is data retrieved through Google search. This method allows us to cover the full range of dimensions, and provides a large and scalable volume of data, but the quality of the content requires further verification. We convert the fine-grained dimensions in the dimensional schema into search keywords and combine them with the target culture to construct retrieval queries(e.g., Celebration of Festivals in Spanish culture). 
In this step, 
we apply GPT-4o to filter the page content, summarize and extract information from the retrieved texts.

\begin{table}[]
    \centering
    \renewcommand{\arraystretch}{1.3} 
    \begin{tabular*}{\columnwidth}{@{\extracolsep{\fill}} c c c c @{}}
    \hline
        \textbf{Culture} & \textbf{Annotator} & \textbf{Knowledge} & \textbf{Evaluation Dataset} \\
        \hline
        \multirow{2}{*}{Spanish} & Annotator1 & 98.57\% & 99.46\% \\
                                 & Annotator2 & 97.85\% & 99.12\% \\
        \multirow{2}{*}{Chinese} & Annotator1 & 97.14\% & 100.0\% \\
                                 & Annotator2 & 99.29\% & 99.11\% \\
    \hline
    \end{tabular*}
    \caption{The quality assessment results of the cultural knowledge and the evaluation dataset via manual sampling inspection. The validation of the cultural knowledge instances focuses on identifying factual or expression-related errors, and the quality of the evaluation set emphasizes the logical consistency among the question, the answer, and the reference knowledge
}
    \label{tab:quality}
\end{table}

\subsection{Evaluation Dataset Generation}

We draw on cognitive science theories\cite{2015Intercultural,Avgousti2018Intercultural,Feng2025Exploring} to define four content types for our questions:  Factual, Conceptual, Misleading and Multi-hop, as this design ensures that the evaluation not only probes different facets of cultural knowledge but also reflects the practical challenges models face in real-world cultural understanding and interaction. These four question types respectively assess whether the model knows cultural facts, understands the underlying meaning of cultural phenomena, can identify cultural biases, and is able to synthesize multiple cultural elements while grasping the deeper logic or internal connections among cultural phenomena. 
The question formats include multiple-choice, true/false, short answer, and essay questions. 
For each type of question, we apply specific expert-designed prompts to clearly define the task requirements. 
Question generation is based on a Retrieval-Augmented Generation (RAG)\cite{lewis2020retrieval} approach, randomly selecting a set of cultural rules for the LLM to reference during generation. Detailed prompt formulations are provided in the Appendix. 

\subsection{Quality Control}
For cultural knowledge and generation-related tasks, the framework provides LLM-based quality evaluation and filtering. 
For the validation of cultural knowledge, we primarily focus on the consistency between the rewritten content and the source webpages, since we found that when using queries such as “eating habits in Spanish culture”, the top five results returned by Google are consistently high-quality webpages that are highly relevant to the target dimension of the query. In addition to widely recognized platforms such as Wikipedia, these results also include official releases from relevant government agencies of the target country, authoritative media outlets, long-established authoritative tourism information portals and so on. 

On the other hand, the evaluation of the generated questions primarily focuses on examining the logical consistency among the question itself, the provided answer, and the underlying reference cultural knowledge. Specifically, it assesses whether the question is formulated in a way that is coherent with the reference knowledge, whether the corresponding answer logically follows from the question, and whether any contradictions or reasoning errors arise across these three components.



\begin{table*}[]
\centering
\resizebox{\textwidth}{!}{
\begin{tabular}{cccccccccccccccc}
\hline

\multirow{2}{*}{\textbf{Models}}  & & & \multicolumn{3}{c}{\textbf{Spanish Culture}} & &  & &  &\multicolumn{3}{c}{\textbf{Chinese Culture}} & & \\
\cmidrule(lr){2-8}
\cmidrule(lr){9-15}

&\textbf{Lang} &\textbf{Acc} & \textbf{G \& C}
& \textbf{PC \& H}
& \textbf{R\& P}
& \textbf{SR \& S}
& \textbf{V \& B}
& \textbf{Lang} & \textbf{Acc} & \textbf{G \& C}
& \textbf{PC \& H}
& \textbf{R\& P}
& \textbf{SR \& S}
& \textbf{V \& B}
\\

\hline 
\rowcolor{gray!10}
\multicolumn{15}{c}{\textbf{General-purpose models}} \\
\hline

\multirow{2}{*}{Qwen2.5-7B} & en  & 0.832 & 0.850 & 0.808 & 0.800 & 0.958 & 0.868
& en  & 0.741  & 0.883 & 0.681 & 0.740 & 0.771 & 0.829\\
                  & sp  & 0.659 & 0.750 & 0.620 & 0.640 & 0.688 & 0.737
                  & cn  & 0.827 & 0.800 & 0.815 & 0.810 & 0.896 & 0.868\\
\multirow{2}{*}{Qwen2.5-14B} & en  & 0.870  & 0.883 & 0.851 & \textbf{\underline{0.840}} & \textbf{\underline{0.958}} & 0.908 
& en  & 0.783  & \textbf{\underline{0.917}} & 0.761 & 0.717 & 0.750 & 0.867 \\
                  & sp  & 0.745 & 0.767 & \textbf{\underline{0.728}} & 0.730 & 0.771 & 0.790 
                  & cn  & \textbf{\underline{0.871}}  & \textbf{\underline{0.917}} & \textbf{\underline{0.844}} & \textbf{\underline{0.880}} & \textbf{\underline{0.896}} & \textbf{\underline{0.908}}
                  \\
\multirow{2}{*}{Qwen2.5-32B} & en & 0.861  & \textbf{\underline{0.900}} & 0.848 & 0.820 & 0.896 & 0.908 
& en & \textbf{\underline{0.798}}  & 0.917 & \textbf{\underline{0.772}} & \textbf{\underline{0.740}} & \textbf{\underline{0.854}} & 0.842 \\
                  
                  & sp & 0.753  & 0.817 & 0.714 & 0.710 & 0.792 & \textbf{\underline{0.880}}
                  & cn & 0.586  & 0.667 & 0.583 & 0.550 & 0.563 & 0.592 \\
\multirow{2}{*}{Llama3-8B} & en & 0.800 & 0.783 & 0.786 & 0.760 & 0.938 & 0.829
& en & 0.651 & 0.683 & 0.627 & 0.576 & 0.792 & 0.720\\
                  & sp & 0.563  & 0.567 & 0.554 & 0.480 & 0.625 & 0.658
                  & cn & 0.655  & 0.610 & 0.634 & 0.600 & 0.729 & 0.790\\

\multirow{2}{*}{GPT-4o} & en & \textbf{\underline{0.875}} & 0.850 & \textbf{\underline{0.870}} & 0.840 & 0.938 & \textbf{\underline{0.921}}
& en & 0.780 & 0.850 & 0.765 & 0.690 & 0.750 & \textbf{\underline{0.921}}\\
                  & sp & \textbf{\underline{0.761}} & \textbf{\underline{0.850}} & 0.710 & \textbf{\underline{0.770}} & \textbf{\underline{0.813}} & 0.829
                  & cn & 0.773  & 0.800 & 0.765 & 0.750 & 0.771 & 0.816\\
                  
\hline
\rowcolor{gray!10}
\multicolumn{15}{c}{\textbf{Deep-reasoning models}} \\
\hline
\multirow{2}{*}{Deepseek-r1} & en & 0.845  & 0.733 & 0.851 & 0.810 & 0.938 & \textbf{\underline{0.895}}
& en & \textbf{\underline{0.837}} & 0.883 & \textbf{\underline{0.801}} & \textbf{\underline{0.840}} & \textbf{\underline{0.854}} & \textbf{\underline{0.921}} \\
                  & sp & 0.664  & 0.810 & 0.629 & 0.586 & 0.729 & 0.737 
                  & cn & 0.870 & \textbf{\underline{0.917}} & 0.848 & 0.840 & \textbf{\underline{0.896}} & \textbf{\underline{0.934}} \\
\multirow{2}{*}{Deepseek-v3} & en &\textbf{\underline{0.879}}  & \textbf{\underline{0.933}} & \textbf{\underline{0.859}} & 0.860 & 0.938 & 0.895
& en & 0.794 & \textbf{\underline{0.900}} & 0.743 & 0.790 & 0.851 & 0.868\\
                  & sp & 0.644  & 0.683 & 0.623 & 0.660 & 0.542 & 0.733 
                  & cn & 0.675  & 0.683 & 0.659 & 0.640 & 0.688 & 0.763 \\

\multirow{2}{*}{Qwen3-8B} & en & 0.834  & 0.833 & 0.786 & \textbf{\underline{0.870}} & \textbf{\underline{0.979}} & 0.868
& en & 0.796 & 0.867 & 0.757 & 0.770 & 0.833 & 0.895\\
                  & sp & \textbf{\underline{0.734}}  & \textbf{\underline{0.817}} & \textbf{\underline{0.703}} & \textbf{\underline{0.720}} & \textbf{\underline{0.792}} & 0.763 
                  & cn & \textbf{\underline{0.893}}  & 0.917 & \textbf{\underline{0.877}} & \textbf{\underline{0.890}} & 0.896 & 0.934 \\
\multirow{2}{*}{Qwen3-8B-f} & en & 0.827  & 0.833 & 0.804 & 0.800 & 0.958 & 0.855
& en & 0.779  & 0.900 & 0.750 & 0.740 & 0.792 & 0.829\\\
                  & sp & 0.713  & 0.817 & 0.688 & 0.680 & 0.688 & \textbf{\underline{0.776}}
                  & cn & 0.838  & 0.867 & 0.823 & 0.810 & 0.875 & 0.882\\
\hline
\rowcolor{gray!10}
\multicolumn{15}{c}{\textbf{Models Continually-trained on Multilingual Corpus}} \\
\hline
\multirow{2}{*}{PolyLM-7B} & en & 0.113  & 0.133 & 0.130 & 0.090 & 0.042 & 0.105
& en & 0.111 & 0.133 & 0.091 & 0.120 & 0.167 & 0.118\\
                  & sp & 0.057  & 0.083 & 0.044 & 0.040 & 0.083 & 0.092 
                  & cn & 0.141 & 0.117 & 0.145 & 0.140 & 0.125 & 0.158 \\
\multirow{2}{*}{PolyLM-13B} & en & 0.458  & 0.367 & 0.491 & 0.390 & 0.500 & 0.474
& en & 0.225 & 0.150 & 0.243 & 0.172 & 0.292 & 0.250\\
                  & sp & 0.204  & 0.200 & 0.207 & 0.250 & 0.063 & 0.224
                  & cn & 0.330 & 0.283 & 0.330 & 0.330 & 0.250 & 0.421\\
\hline
\end{tabular}
}

\caption{Main evaluation results on Spanish culture and Chinese culture. We evaluate the question-answer accuracy of several types of LLMs on different languages and cultural categories mentioned above. Lang refers to the language of the questions. Qwen3-8B-f refers to Qwen3-8B without deep reasoning enabled. Abbreviations follow a consistent pattern: for example, “G\&C” stands for Geography and Customs; other abbreviations follow the same convention.}
    \label{tab:MainResults}
\end{table*}
\section{Evaluation Benchmark}
\subsection{Dataset Construction}
Leveraging the proposed framework, we independently construct evaluation datasets grounded in Chinese culture and Spanish-speaking cultural contexts. For each culture, we conduct knowledge extraction and evaluation dataset generation in both English and the corresponding native language (i.e., Spanish or Chinese). We conduct a categorical analysis of the source websites of the cultural knowledge instances. 
The distribution of webpage types is shown in Appendix. We find that online cultural data is inherently heterogeneous, with a broad spectrum of sources that complement each other. 
In addition, we cluster the extracted culture knowledge instances, obtaining \textbf{1.1k} cultural keywords to further enrich the dimensional schema.

\subsection{Evaluation Metric}


We adopt an automated evaluation method to analyze and summarize results on the dataset, using answer accuracy as the primary metric, defined as the proportion of correct responses over all evaluated items. For objective questions (e.g., multiple-choice, true/false), we apply direct answer matching: a prediction $\hat{y}$ is marked correct if it matches the ground truth $y$. For subjective questions(e.g., short answers, essays), we adopt the LLM-as-a-Judge approach\cite{Desmond2025EvalAssist}, where the question $q$, reference answer $a_{\text{ref}}$, and model-generated answer $a_{\text{test}}$ are jointly evaluated by a Judge Model to determine whether $a_{\text{test}}$ aligns with the key semantic content of $a_{\text{ref}}$. 

Due to the flexibility of answers in subjective questions, we require the Judge Model to compare the reference answer with the model-generated answer and determine whether the content of the latter conflicts with the cultural knowledge in the reference. If a conflict exists, the answer is considered incorrect. This approach ensures that the evaluation captures cultural understanding while allowing a certain degree of answer flexibility.

\begin{equation}
\text{Acc} =
\frac{1}{N} \sum_{i=1}^{N} 
\mathbf{1}\Big(
\begin{cases}
\hat{y}_i = y_i, & \text{Obj} \\
\text{Judge}(q_i, a_i^{\text{ref}}, a_i^{\text{test}}), & \text{Subj}
\end{cases}
\Big)
\end{equation}

\subsection{Statistics}
Our constructed benchmark consists of \textbf{11962} cultural knowledge instances and \textbf{44.1k} questions in total, Table~\ref{tab:statistics} shows the detailed statistics of each culture and language. The distribution of the English version of the cultural knowledge instances is illustrated in Figure~\ref{fig:databaseDistribution}. The distribution of cultural knowledge instances exhibits a trend similar to that of the dimensions in the schema, a pattern that holds consistently across different languages. 

We invite human experts to manually evaluate the quality of the culture knowledge and the correctness of the evaluation dataset. 
Table~\ref{tab:quality} shows the quality evaluation results of the constructed benchmark. 
The validation of cultural knowledge focuses on identifying factual or expression-related errors, whereas the evaluation of the dataset questions emphasizes the logical consistency among the question, the answer, and the reference knowledge. 
The results show that the generated cultural knowledge and evaluation questions are of high quality and can be used to assess the cultural understanding capabilities of LLMs, which also validates the effectiveness of our automatic quality control method.

\section{Experiments}
We carry out a series of experiments to examine: 1) To what extent can current LLMs align with specific cultural knowledge. 2) What are the factors that influence the performances of LLMs on culture-specific knowledge-related tasks. 3) Whether there is a noticeable difference in the performances of LLMs across different cultural dimensions, languages, and types of question.


\paragraph{Models and Implementation}
Our evaluation include three types of LLMs: General-purpose models, including Qwen2.5-7B-Instruct, Qwen2.5-14B-Instruct, Qwen2.5-32B-Instruct\cite{qwen2.5_2024} and GPT-4o\cite{hurst2024gpt}; Deep-reasoning models and its non-deep-reasoning version, including Qwen-3\cite{yang2025qwen3}, Deepseek-r1\cite{guo2025deepseek}and Deepseek-v3\cite{liu2024deepseek}; models trained specifically on multilingual data, including polylm-qwen-7b and polylm-qwen-13b\cite{wei2023polylm}.
All experiments are conducted using NVIDIA A800 GPUs with 80GB memory. Local models are loaded using FP16 precision and implemented with PyTorch and the Transformers library. The generation limit is 1024 new tokens.

\subsection{Results}


The overall performances of all LLMs are listed in Table~\ref{tab:MainResults}.
We evaluate different models across all dimensions, languages, and question types, and analyzed potential influencing factors. 

\textbf{Observation 1: The cultural understanding ability of LLMs is language-dependent.}
Overall, as is shown in Table~\ref{tab:MainResults}, 
the accuracy of model responses to cultural questions is directly related to the language of the questions. 
For Spanish-speaking cultures, When the language switch from English to Spanish, all models show a trend of accuracy decline. However, for Chinese culture, many models perform better on questions in Chinese than in English. 
This indicate that there is still a performance gap between current LLMs in English and other relatively low-resource languages. 

\begin{figure}[t]
\centering
  \includegraphics[scale=0.38]{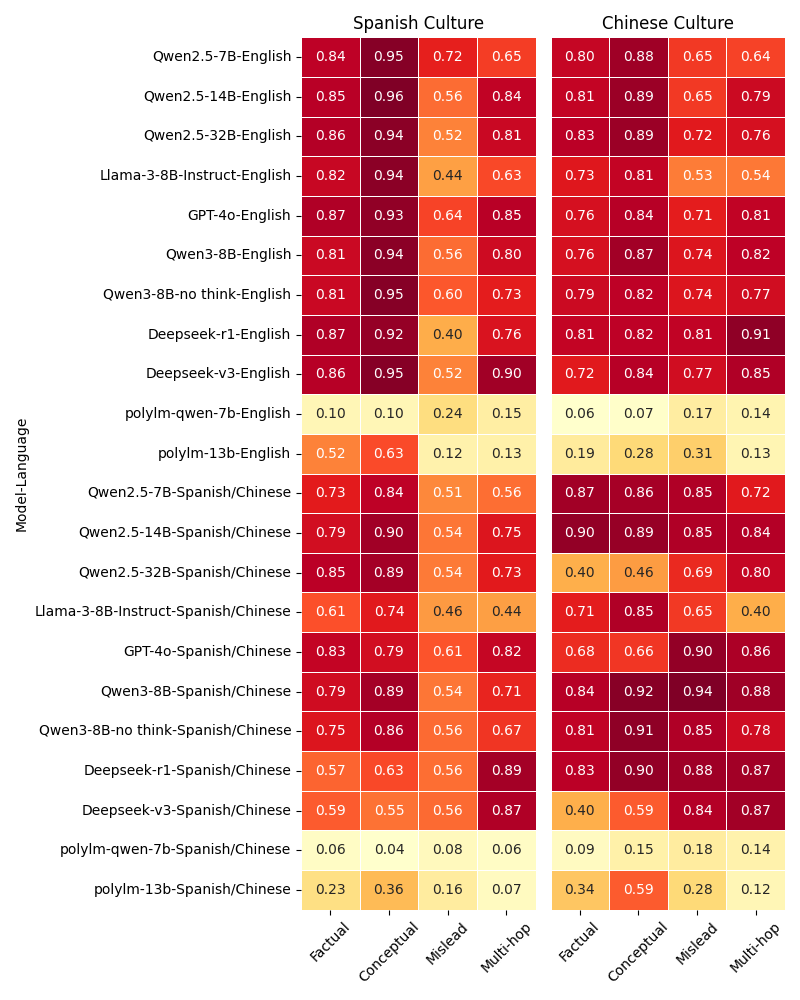}
  \caption{Models' performances across different question content types. \textit{Factual} evaluates the understanding of factual cultural knowledge; \textit{Conceptual} assesses the models' grasp of the deeper meanings of cultural concepts; Mislead examines whether the models can detect cultural bias; and Multi-hop evaluates the models' ability to apply cultural knowledge in real-world communicative scenarios.}
  \label{fig: heat_type}
\end{figure}

\begin{figure}[t]
\centering
  \includegraphics[scale=0.30]{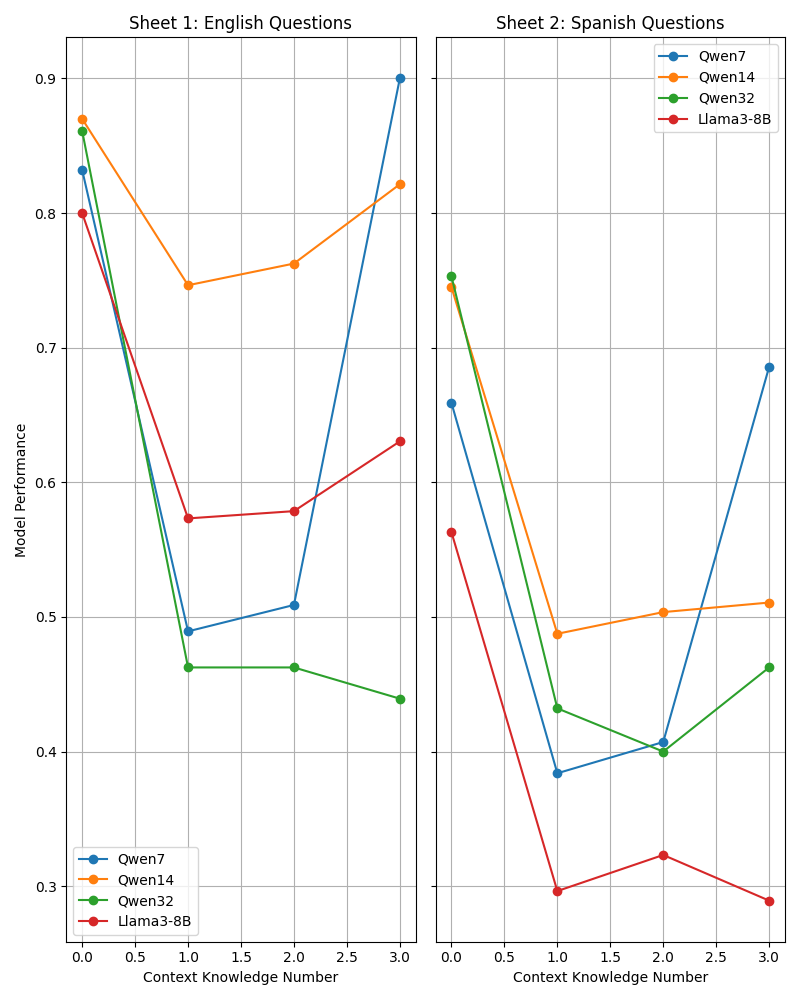}
  \caption{Different models' performance variation with different amounts of injected cultural knowledge via prompts. }
  \label{fig: rag}
\end{figure}
On the other hand, the models demonstrate \textbf{a culture-specific weakness in identifying cultural bias}. As is shown in Figure~\ref{fig: heat_type}, the models' performances on handling Mislead-type questions are significantly lower than on other types within Spanish culture, while in Chinese culture, such a disparity is less pronounced and exhibits lower consistency across models. This suggests that the models may be more susceptible to misleading content or less robust in reasoning under deception in specific cultural contexts, highlighting potential cultural sensitivity or bias in their reasoning capabilities.

\begin{figure*}[!t]
  \centering

  \begin{subfigure}[t]{0.48\textwidth}
    
    \includegraphics[height=0.34\textheight, width=\textwidth]{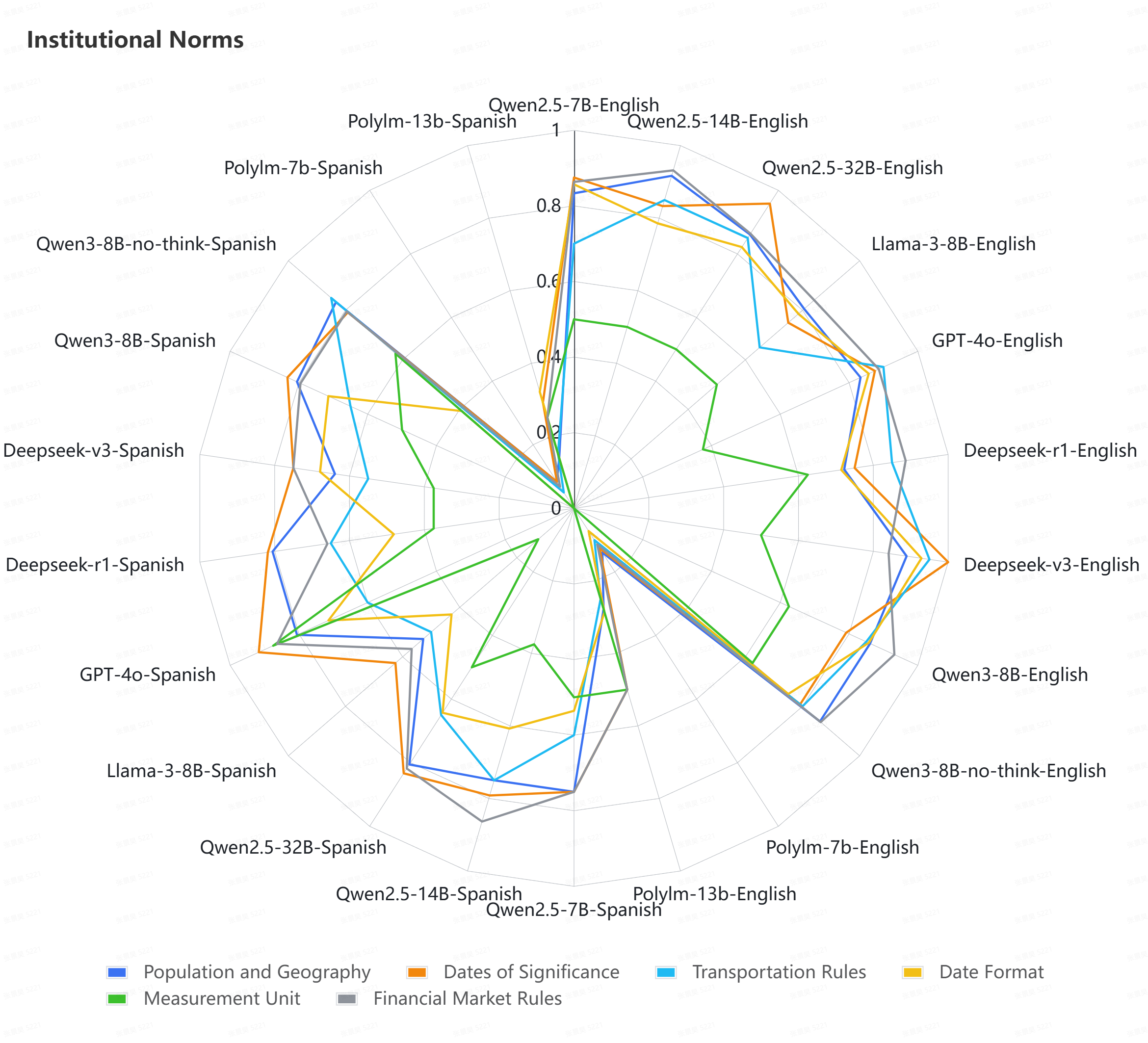}

  \end{subfigure}
  \hfill
  \begin{subfigure}[b]{0.48\textwidth}
    \includegraphics[height=0.34\textheight, width=\textwidth]{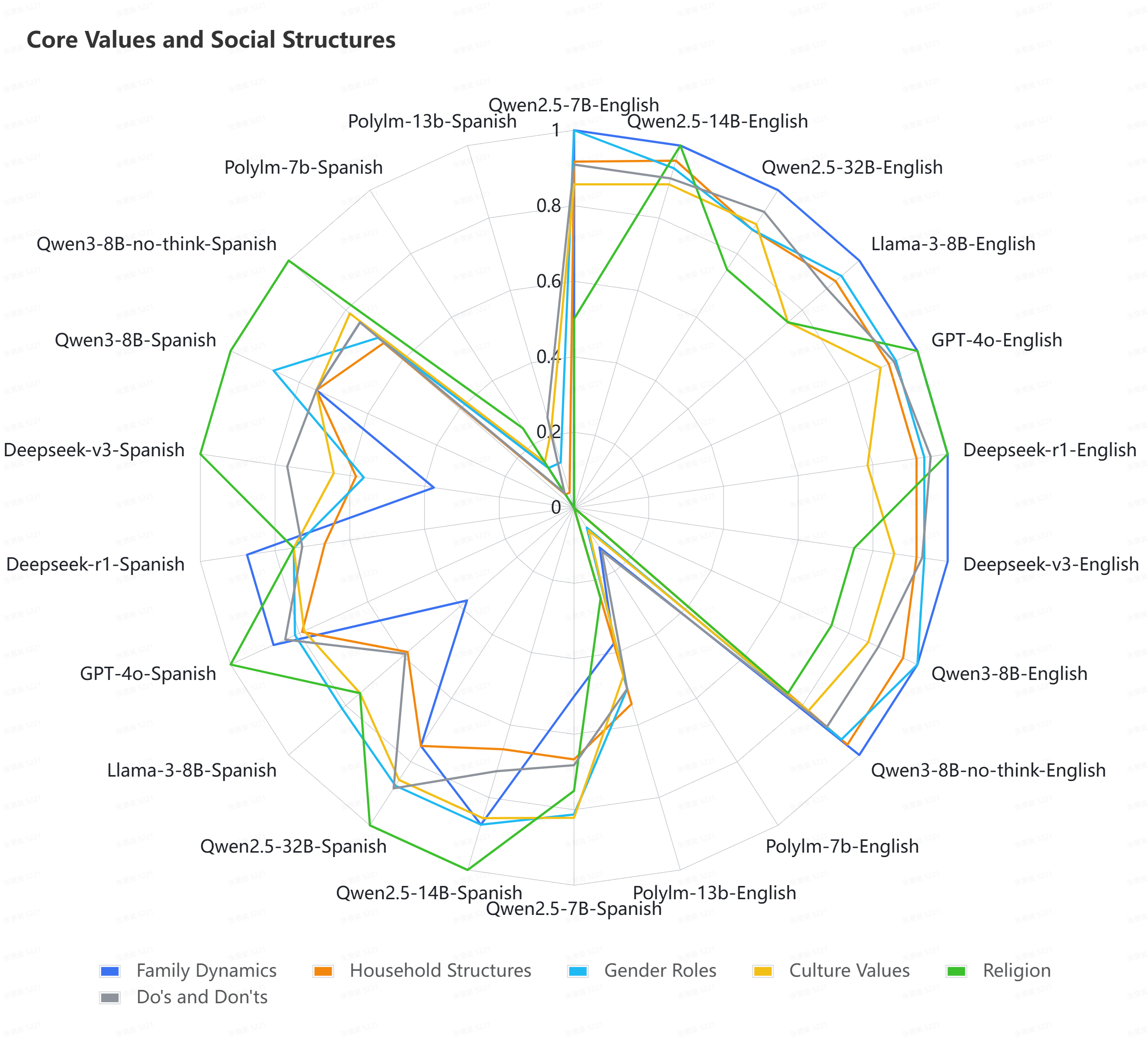}
    \end{subfigure}





  \caption{Performances of models on the Institutional Norms Layer and Core Values and Social Structures Layer, within the Spanish cultural context. For institutional norms, performances across dimensions is similar in both English and Spanish. However, for values and social structures, many dimensions exhibit significant differences between the two languages, such as \textit{Family Dynamics} and \textit{Religion}.
}
  \label{fig:all_layers}
\end{figure*}

Meanwhile, all models exhibit \textbf{varying performances across different \textit{Topic Aspects}}. 
We evaluate the finer-grained \textit{Categories} within each \textit{Layer} and present a subset of the results in the Figure~\ref{fig:all_layers}, and the complete results can be found in the Appendix. 
In the context of Spanish culture, as the cultural knowledge evaluation progresses from shallow to deep layers, the models exhibit increasing performance divergence within the same dimension when questioned in different languages. 
At the \textbf{Institutional Norm Layer}, 
all models perform poorly on questions related to \textit{Measurement Units}, while demonstrating better mastery of knowledge concerning \textit{Dates of Significance}. 
At the \textbf{Core Values and Social Structures Layer}, for \textit{Family Dynamics} and \textit{Household Structures}, almost all models answer English questions very well but perform poorly on Spanish ones. 
For Chinese culture, however, as the evaluation deepens, the models’ performances across different languages tend to stabilize within the same dimension.




\textbf{Observation 2: Larger models typically exhibit stronger performance on cultural understanding tasks}, as they encode a broader range of knowledge.
Models from the same series exhibit a generally positive correlation between performance and \textbf{model size}. On one hand, for Spanish culture, increasing model size consistently leads to notable performance gains across all models on Spanish-language questions. 
On the other hand, for Chinese culture, as the scale increases, nearly all models show improved performance on both English and Chinese questions.

\textbf{Observation 3: Deep reasoning does not inherently yield substantial gains on cultural understanding capability}
When a model does not possess adequate knowledge of the relevant culture, deeper reasoning processes may not lead to better performance on cultural questions.
Although models with deep thinking enabled outperform those without it in the context of Chinese culture, a similar advantage is not observed for Spanish-speaking cultures. 
Deep thinking models we used(Deepseek-r1 and Qwen3) were trained on larger Chinese corpora and thus possess stronger Chinese language capabilities. This suggests that deep thinking yields greater benefits only when the model has been trained on sufficiently rich corpora in the target language. Without access to relevant knowledge, no amount of reasoning can fully compensate for the limitations in cultural understanding.

\textbf{Observation 4: Language$\neq$Culture.}
As shown in the Figure~\ref{fig:all_layers}, compared to the similarly sized Qwen models, PolyLM\cite{wei2023polylm}, which incorporates additional multilingual data, exhibits a significant performance drop for both Spanish and Chinese cultures, rather than an improvement. This indicates that multilingualism does not equate to multicultural understanding, and increasing the proportion of data in a particular language during training does not necessarily enhance cultural capability evaluation.

\textbf{Observation 5: Incomplete external cultural knowledge can impair the model's performance on cultural understanding tasks}. Figure~\ref{fig: rag} shows the models’ performance on the evaluation dataset when injecting 0, 1, 2, and 3 pieces of cultural knowledge instances via prompts. We observe that when only a small amount of external knowledge is injected, the model’s performance is even worse than reasoning solely with its internal knowledge. Only when a sufficient amount of relevant external cultural knowledge is provided does a model’s performance improve, and the amount of cultural knowledge required for performance gains varies across different models.

\section{Conclusion}
In this study, we address the critical challenge of systematically and comprehensively evaluating the cultural understanding capability of LLMs. We propose a theoretically grounded and extensible cultural knowledge dimensional framework that satisfies four core principles: comprehensiveness, operationalizability, interpretability, and scalability. In this framework, we develop an automated pipeline to extract high-quality culture knowledge instances and evaluation datasets across any given languages and cultures. Our experimental results on Spanish and Chinese cultures reveal that models exhibit significant differences in cultural capability across languages and dimensions. We find that cultural capability fundamentally depends on a model’s cultural knowledge rather than on language knowledge alone, and that deep reasoning does not reliably compensate for cultural knowledge gaps. These findings highlight the limitations of conventional multilingual training for achieving true cultural alignment,  
which offers valuable insights for the future training, evaluation, and deployment of culturally aligned LLMs.

\section{Ethical Statement}
This work involves the construction and use of a large-scale cultural understanding evaluation dataset, which contains content ranging from common cultural norms to sensitive aspects such as values and beliefs. The data was collected exclusively from publicly available internet sources, including educational materials, open-access cultural databases, and publicly shared questionnaires or forums. No private, proprietary, or identifiable personal data was included.
We acknowledge that cultural data can reflect deeply held values and potentially contentious viewpoints. In constructing the dataset, we took care to avoid harmful stereotypes and minimizing cultural bias. When selecting and annotating the data, we consulted domain experts in cultural studies to ensure contextual fidelity and respect toward the represented communities.
Furthermore, the dataset is intended solely for research purposes, such as evaluating the capabilities of language models in understanding cultural knowledge. It is not designed to make normative claims about cultures, nor should it be used for profiling, ranking, or stereotyping individuals or groups.
To mitigate risks of misuse, we restrict the release of the dataset under a research-use license and require users to agree not to employ it for downstream applications that could cause harm, including but not limited to cultural discrimination, misinformation propagation, or automated decision-making in sensitive contexts (e.g., hiring, immigration, education).
We are committed to ongoing ethical reflection and welcome feedback from cultural communities, stakeholders, and the wider research community.

\bibliography{main}

\section{Supplementary Data}
\paragraph{Experimental Results}
We evaluate the finer-grained \textit{Categories} within each \textit{Layer} and the results are shown in Figure~\ref{fig:all_layers}. 
The use of different languages leads to notable differences in cultural understanding capability across various dimensions. 
We find that the models perform differently across different
Layers depending on the cultural context. 

In the context of Spanish culture, as the cultural knowledge evaluation progresses from shallow to deep layers, the model exhibits increasing performance divergence within
the same dimension when questioned in different languages.
At the Institutional Norm layer, evaluation results across different languages show a certain degree of consistency. All models perform poorly on questions related to Measurement Units, while demonstrating better mastery of knowledge concerning Dates of Significance. At the Behavioral Habits layer, models exhibit different preferences across languages. For Art and Entertainment questions in English, all models perform poorly, whereas on Spanish questions, Art and Entertainment outperforms other secondary dimensions. At the Cultural Values layer, language-dependent preference differences become more pronounced. Nearly all models perform well on Religion-related Spanish questions but poorly on the same questions in English. Conversely,
for Family Dynamics and Household Structures, almost all models answer English questions very well but perform poorly on Spanish ones.

In contrast, for Chinese culture, as the evaluation deepens, the models’ performance across different languages tend to stabilize within the same dimension.

\begin{figure*}

  \begin{subfigure}[t]{0.48\textwidth}
    \includegraphics[height=0.30\textheight, width=\textwidth]{Institutional-Norms-sp.png}
    \caption{Institutional Norms of Spanish culture}
  \end{subfigure}
  \hfill
  \begin{subfigure}[t]{0.48\textwidth}
    \includegraphics[height=0.30\textheight, width=\textwidth]{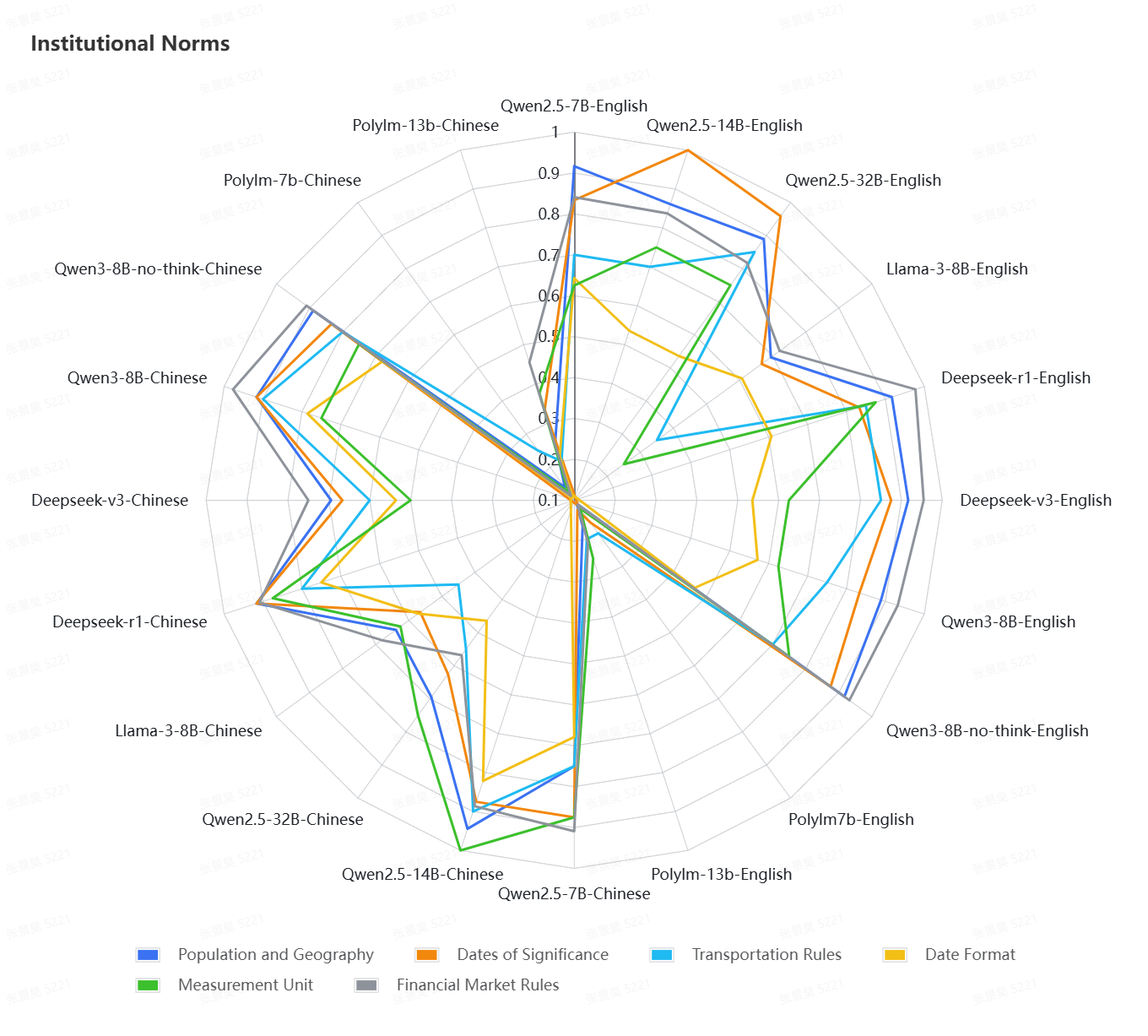}
    \caption{Institutional Norms of Chinese culture}
  \end{subfigure}

  \vspace{0.2em}

  \begin{subfigure}[t]{0.48\textwidth}
    \includegraphics[height=0.30\textheight, width=\textwidth]{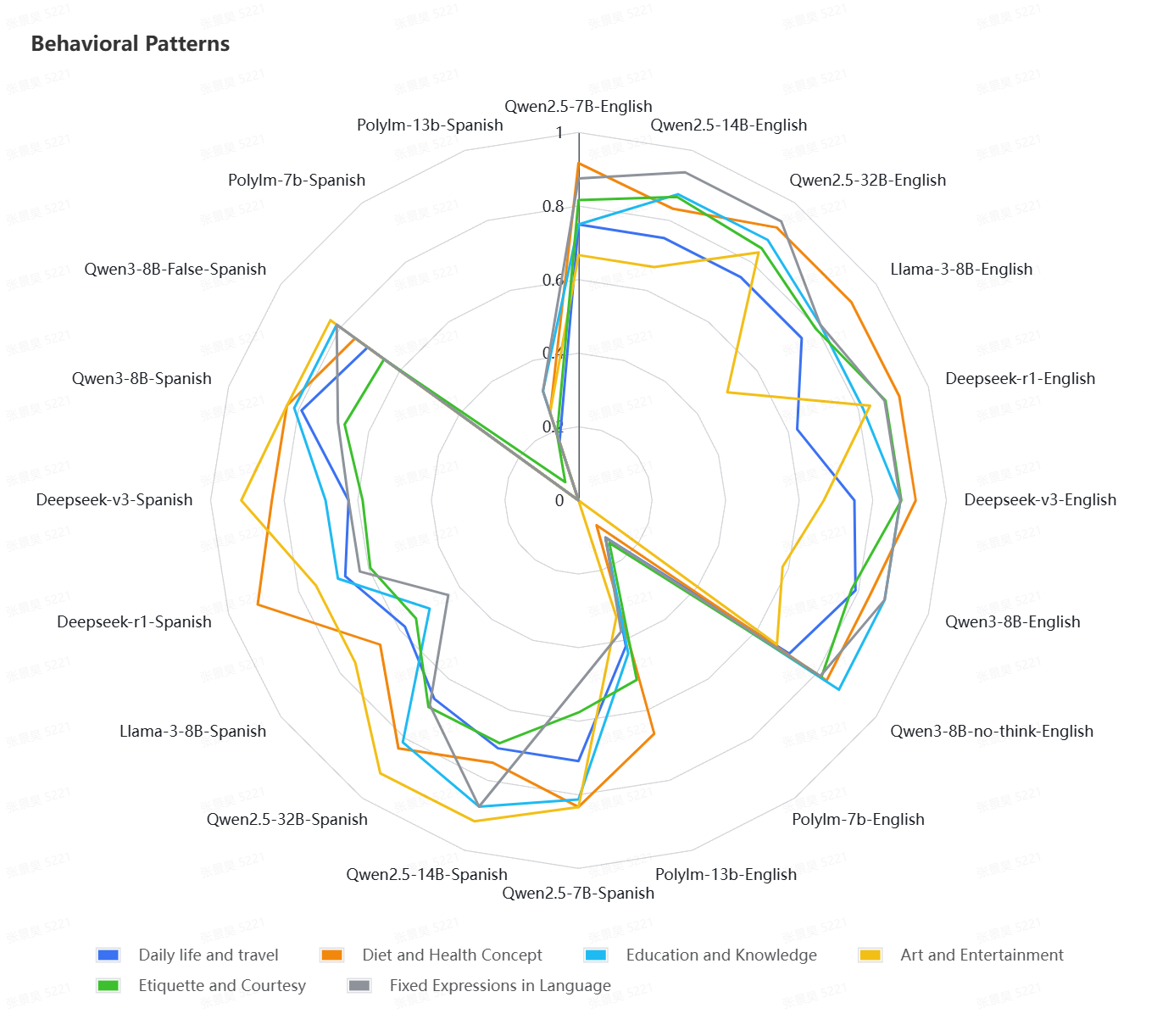}
    \caption{Behavioral Patterns of Spanish culture}
  \end{subfigure}
  \hfill
  \begin{subfigure}[t]{0.48\textwidth}
    \includegraphics[height=0.30\textheight, width=\textwidth]{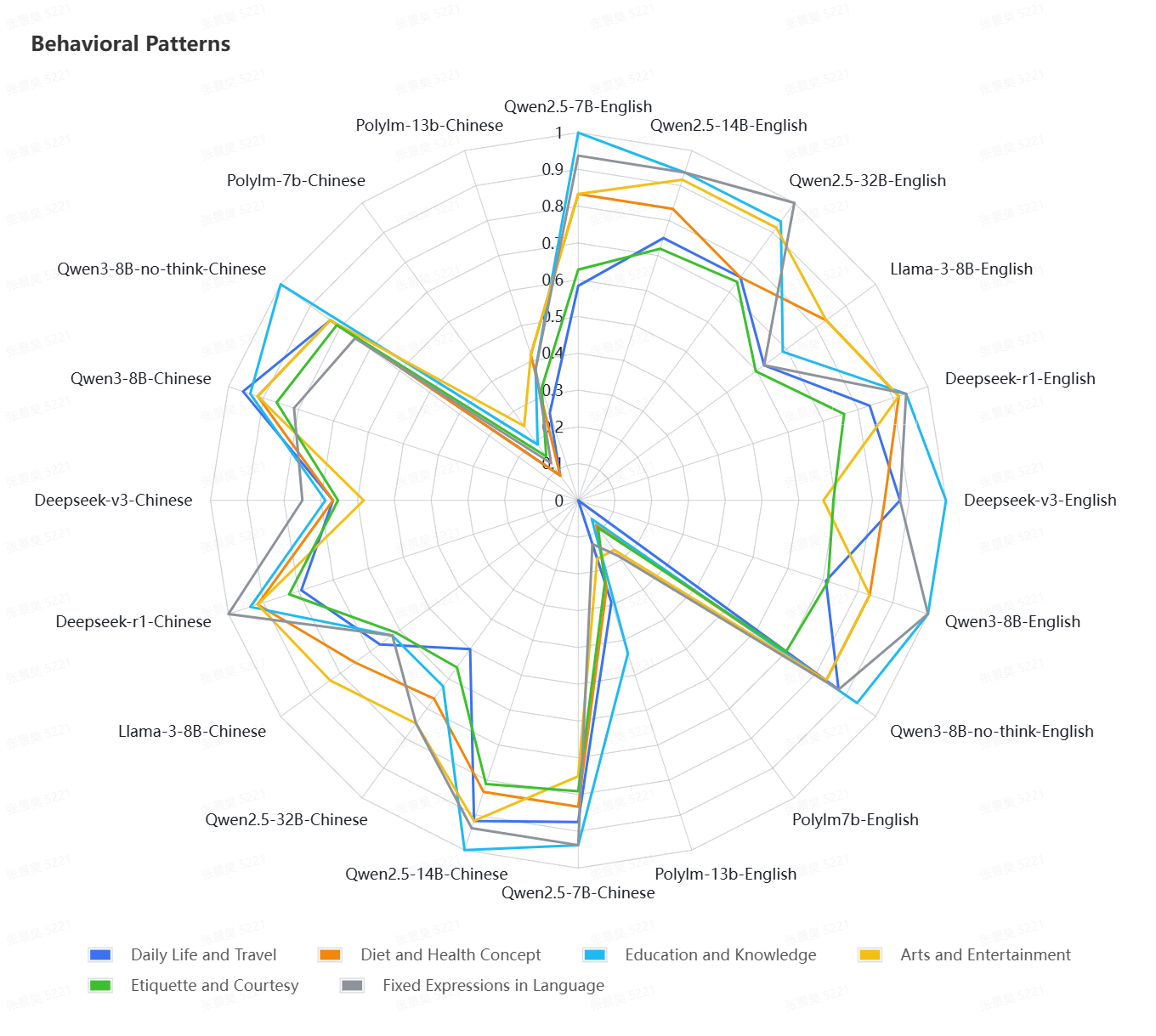}
    \caption{Behavioral Patterns of Chinese culture}
  \end{subfigure}

  \vspace{0.2em}

  \begin{subfigure}[t]{0.48\textwidth}
    \includegraphics[height=0.30\textheight, width=\textwidth]{Core-Values-and-Social-Structures-sp.png}
    \caption{Core Values and Social Structures of Spanish culture}
  \end{subfigure}
  \hfill
  \begin{subfigure}[t]{0.48\textwidth}
    \includegraphics[height=0.30\textheight, width=\textwidth]{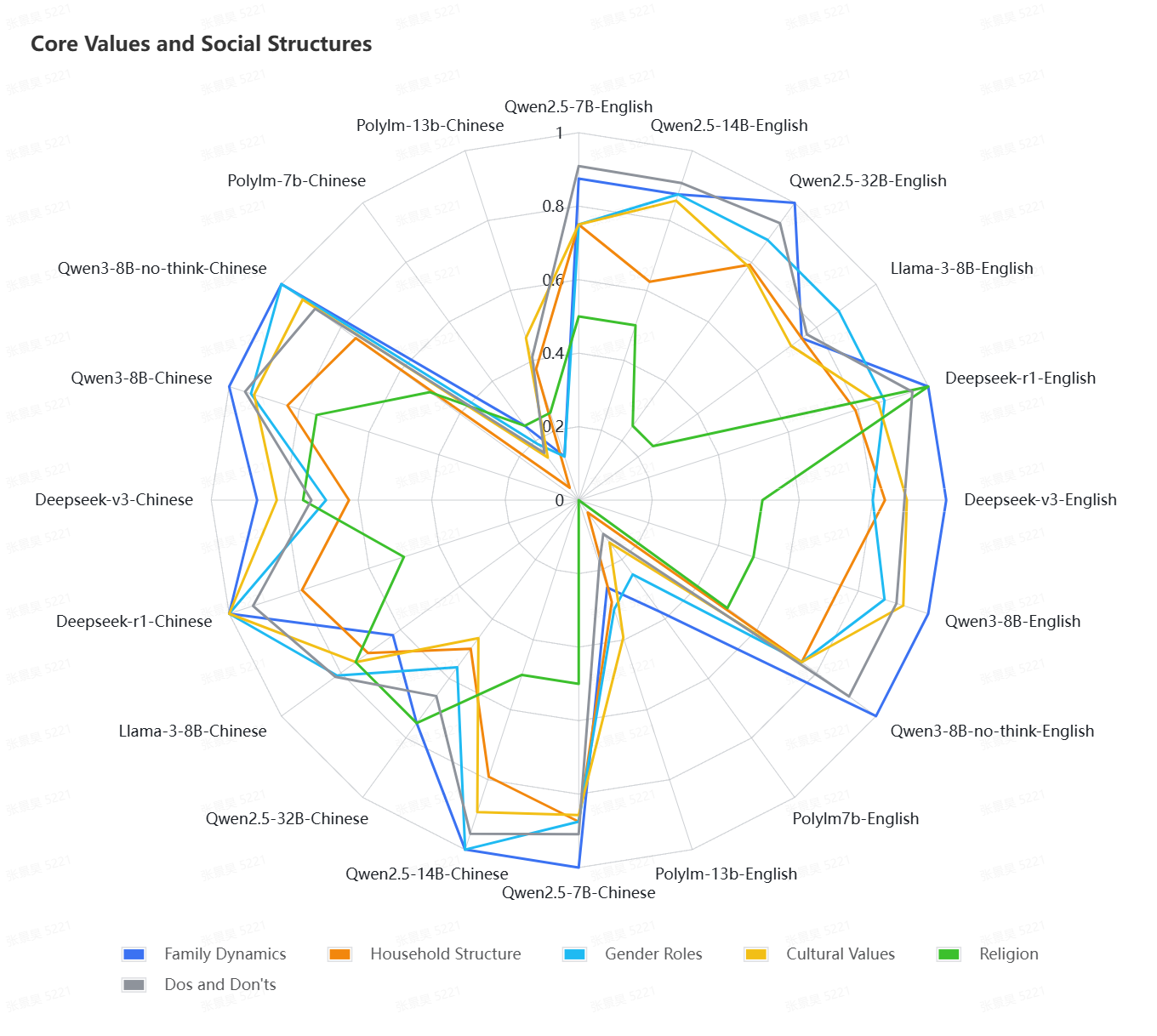}
    \caption{Core Values and Social Structures of Chinese culture}
  \end{subfigure}

  \vspace{0.2em}

  \caption{Performance of Models across Layers and Languages}
  \label{fig:all_layers}

\end{figure*}

\begin{figure*}[!t]
  \centering
  \begin{subfigure}[b]{0.45\textwidth}
    \centering
    \includegraphics[height=0.31\textheight, width=\textwidth]{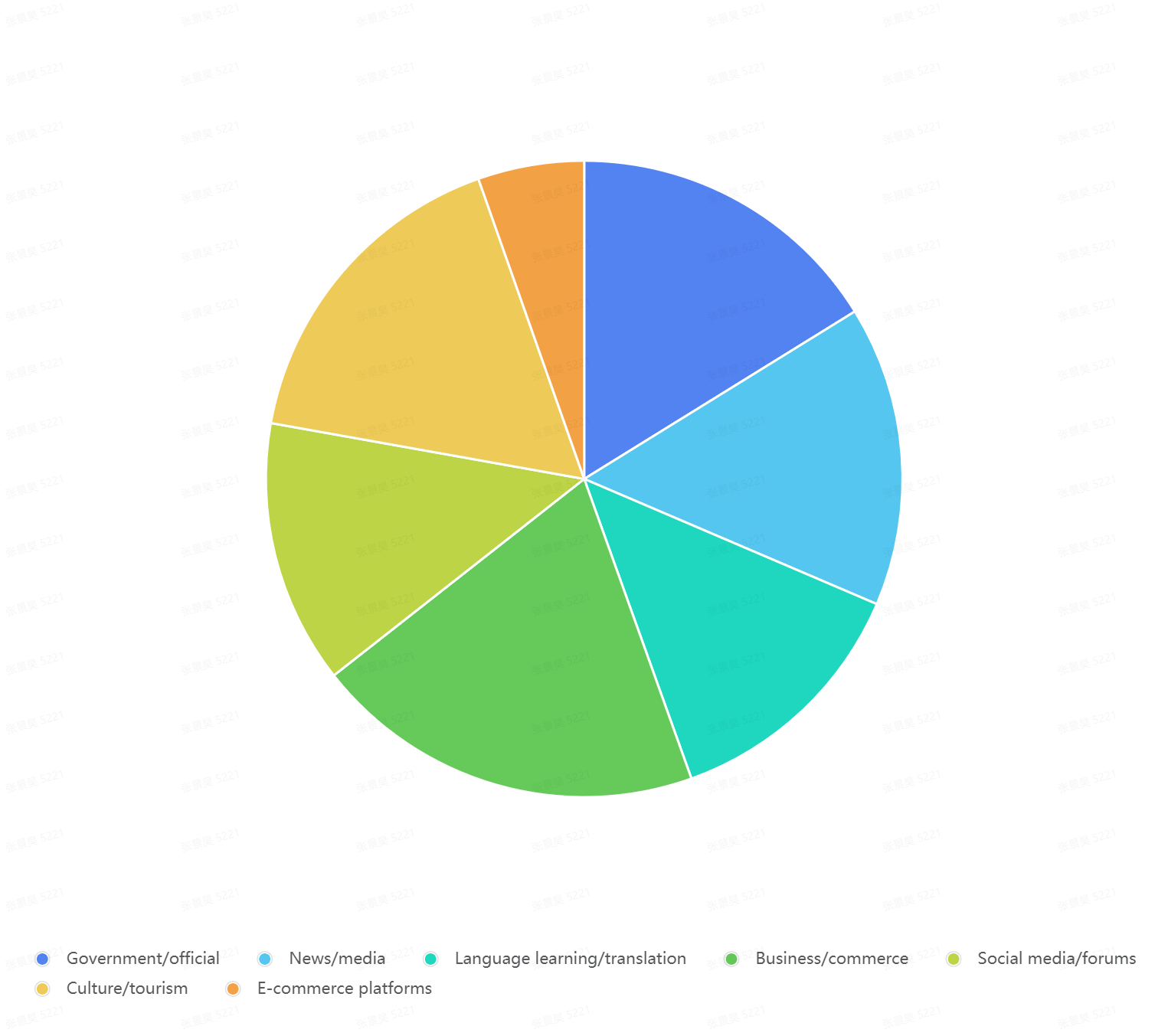}
    \caption{Spanish Culture}
    \label{fig:EN}
  \end{subfigure}
  \hfill
  \begin{subfigure}[b]{0.45\textwidth}
    \centering
    \includegraphics[height=0.31\textheight, width=\textwidth]{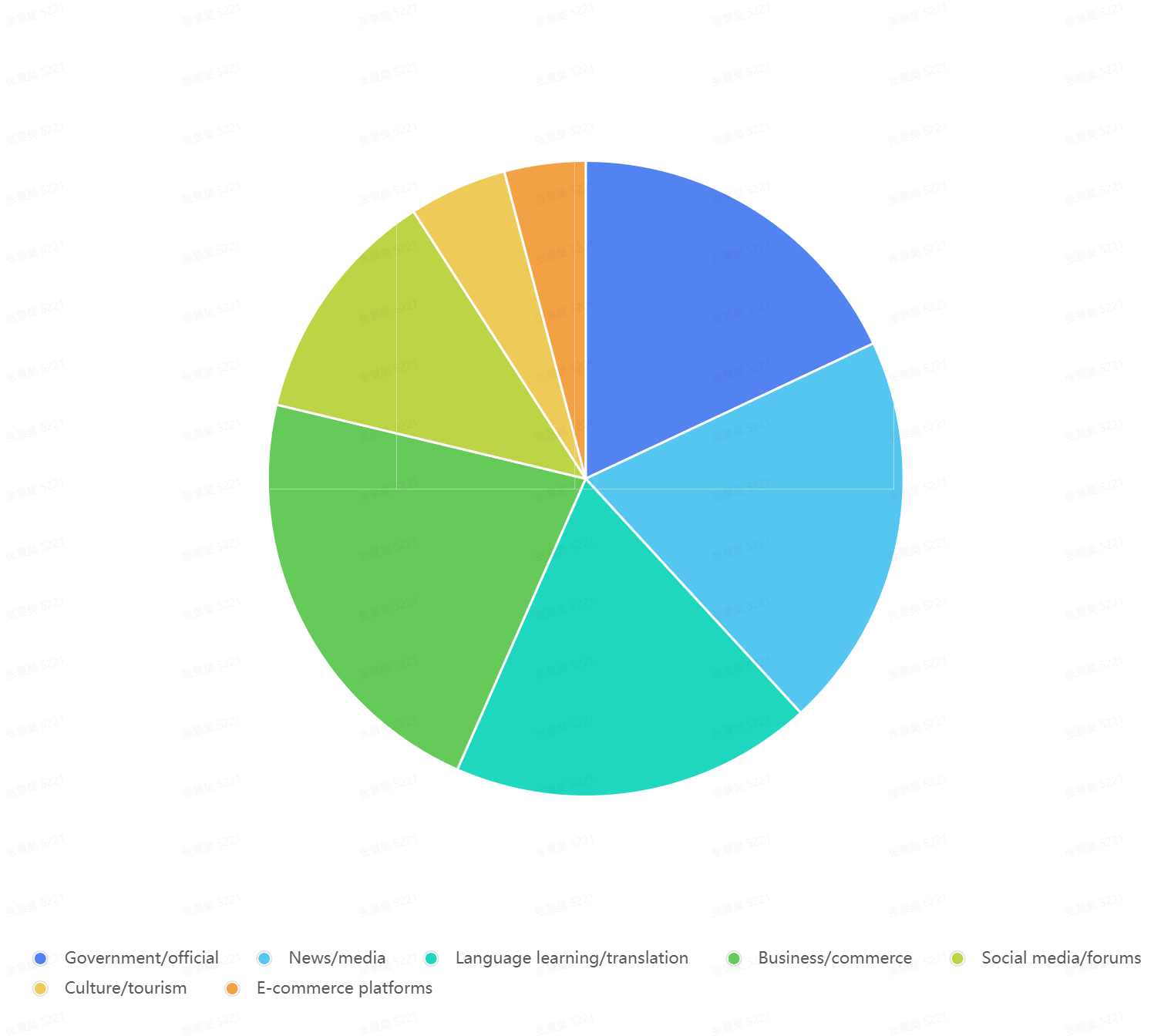}
    \caption{Chinese Culture}
    \label{fig:SP}
  \end{subfigure}

  \caption{Data Source Statistics for Different Extracted Cultural Knowledge Instances}
  \label{fig:datasource}
\end{figure*}

\paragraph{Statistics}In addition to the statistics mentioned in the main technical content, we also analyze the data distribution and sources of the extracted cultural knowledge instances. Figure~\ref{fig:datasource} shows the statistics of data source for Spanish and Chinese culture bases retrieved in corresponding languages. The data sources of the two knowledge bases exhibit distinct distributions: approximately 20\% of the Spanish knowledge base is derived from cultural and tourism websites, whereas this source category accounts for only a small proportion in the Chinese knowledge base. This highlights certain differences in the distribution of internet data across languages within cultural contexts. 

\begin{figure*}[b]
    \centering
  \begin{tabular}[b]{cc}
    \begin{subfigure}[b]{0.4\textwidth}
      \includegraphics[height=0.26\textheight, width=\textwidth]{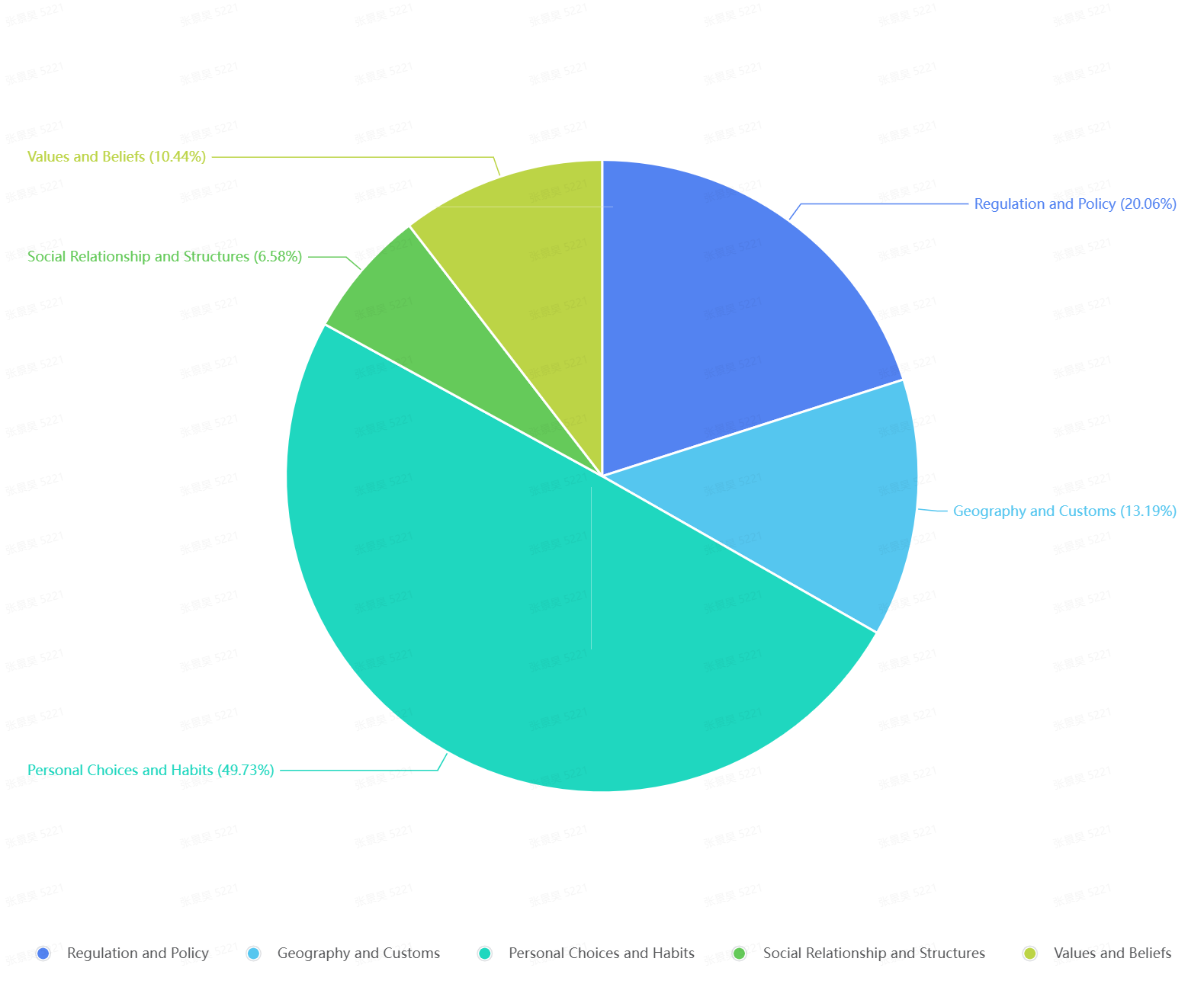}
      \caption{Spanish Cultural Knowledge in English}
    \end{subfigure}
    \hfill
    \begin{subfigure}[b]{0.4\textwidth}
      \includegraphics[height=0.26\textheight, width=\textwidth]{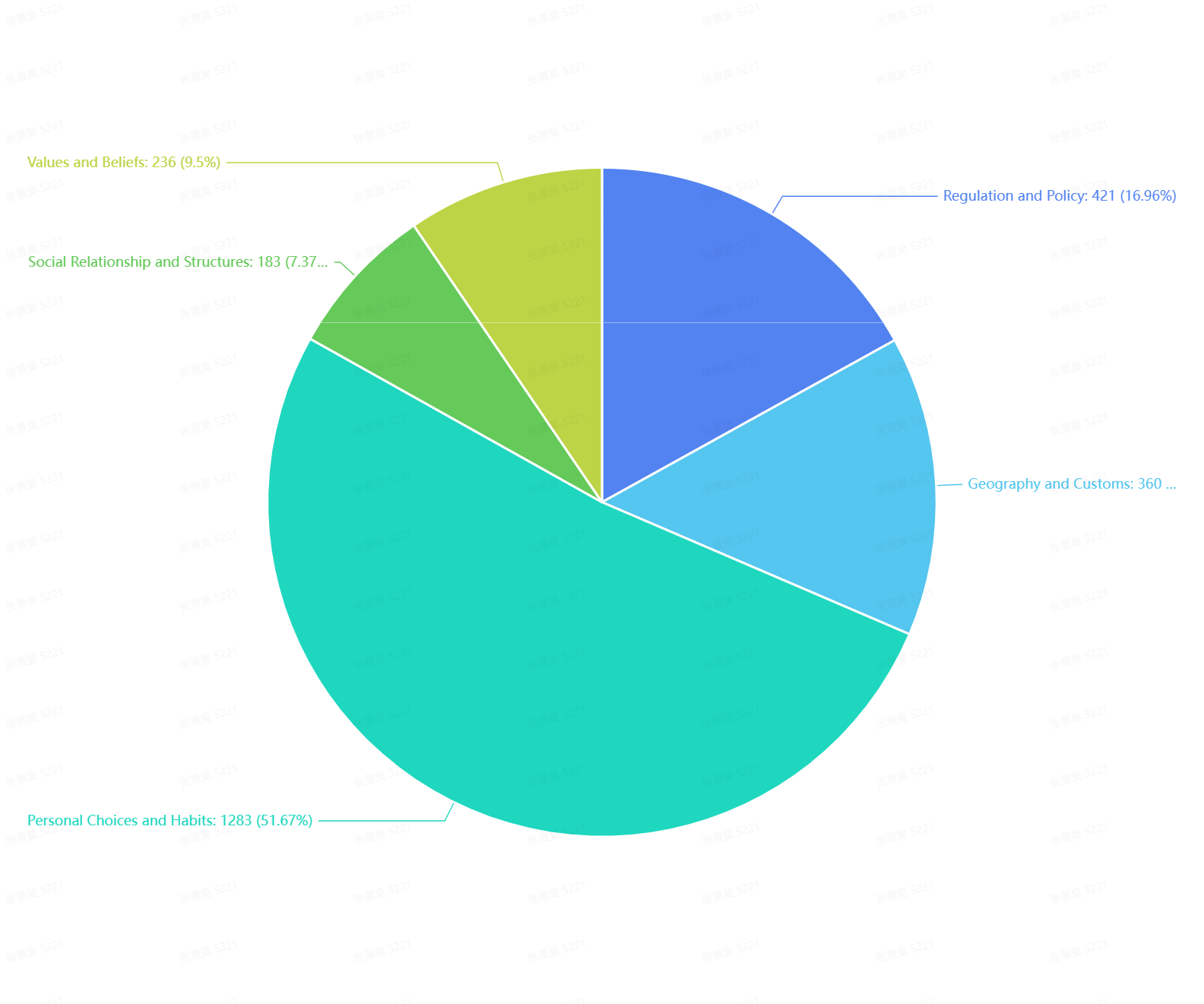}
      \caption{Spanish Cultural Knowledge in Spanish}
    \end{subfigure}
    
    \\
    
    \begin{subfigure}[b]{0.4\textwidth}
      \includegraphics[height=0.26\textheight, width=\textwidth]{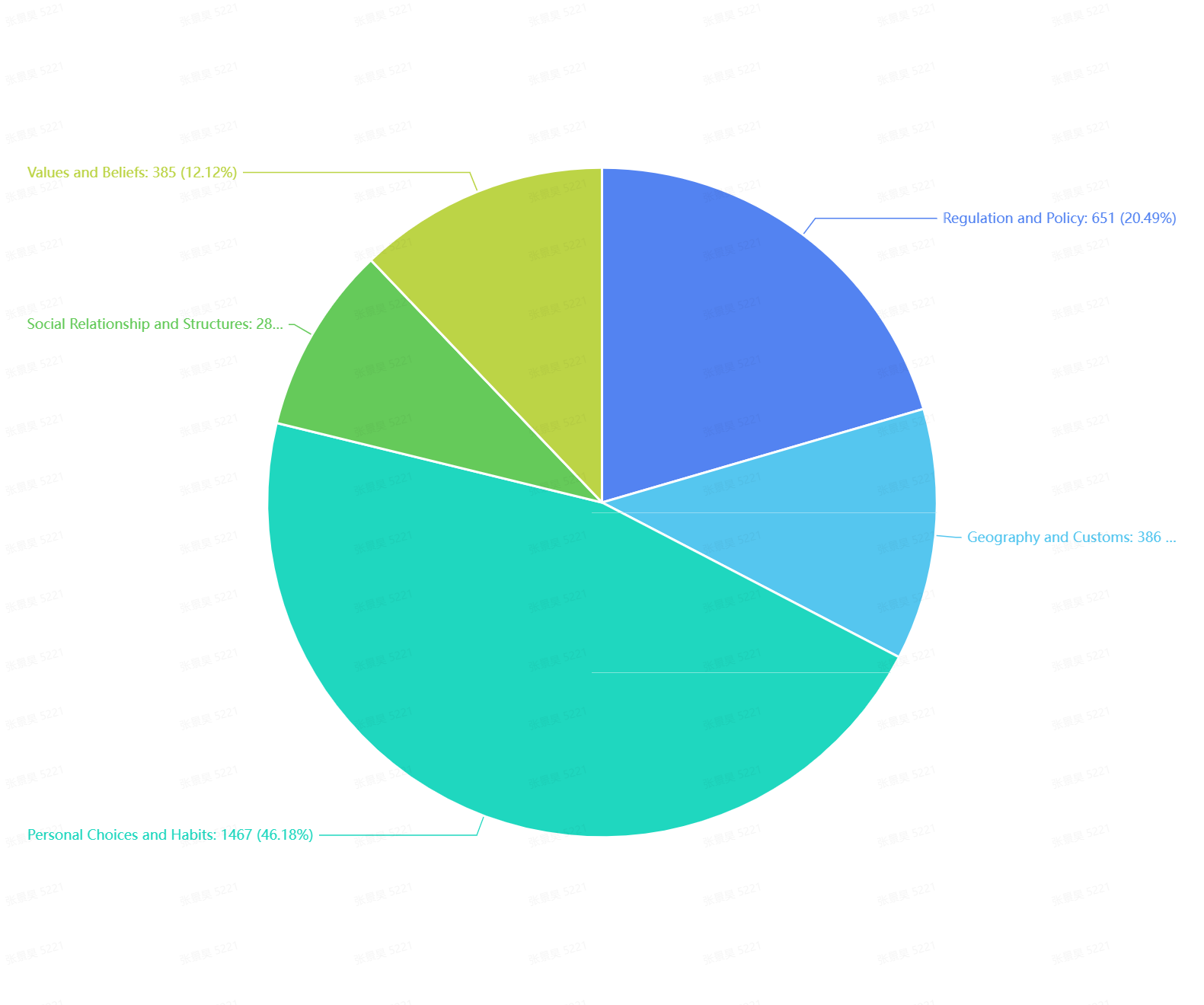}
      \caption{Chinese Cultural Knowledge in English}
    \end{subfigure}
    \hfill
    \begin{subfigure}[b]{0.4\textwidth}
      \includegraphics[height=0.26\textheight, width=\textwidth]{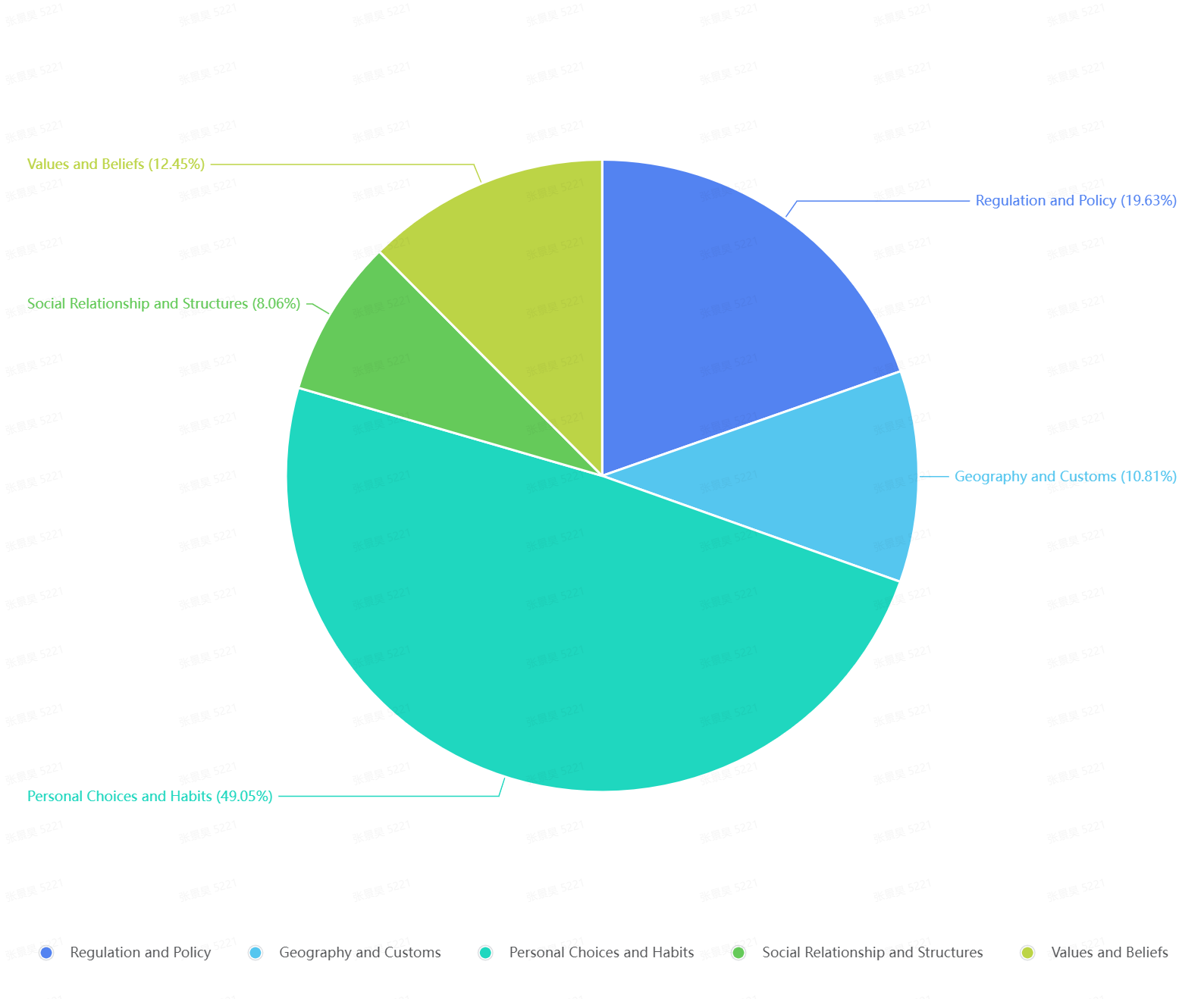}
      \caption{Chinese Cultural Knowledge in Chinese}
    \end{subfigure}
  \end{tabular}
  
  \caption{The distribution of knowledge instances retrieved through searches in different languages.}
  \label{fig:distribution}
\end{figure*}

Figure~\ref{fig:distribution} illustrates the distribution of cultural knowledge across dimensions in different cultural knowledge bases retrieved using different languages. Despite notable differences in data sources among the knowledge bases, the knowledge distributions appear to be largely consistent, following similar patterns that align with the number of dimensions defined in the taxonomy. This suggests that the density of culturally relevant textual content on the Internet is relatively uniform across languages, and that through retrieval and rewriting, comparable amounts of cultural knowledge can be obtained for each dimension.

\section{Detailed Dimensions}
Our proposed dimension schema contains 4 levels, \textit{Cultural Layers, Category, Topic Aspect and Fine-grained Dimensions}. There are in total 3 Cultural Layers, 5 Categories, 18 Topic Aspects and 140 Fine-grained Dimensions, encompassing a wide range of cultural knowledge from explicit commonsense facts, behavioral patterns to implicit values. We transfer the most detailed dimensions into keywords for Google search, thereby effectively guiding the construction and evaluation of the cultural knowledge base. A detailed description of the dimensions is provided in Table~\ref{tab:dims}.

\begin{table*}[t]
  \centering
  \begin{tabularx}{\textwidth}{|l|X|l|}
   \hline
   Dimension & Knowledge & Clustered category\\
   \hline
    \multirow{5}{*}{Alcohol} & [1] Alcohol is deeply integrated into daily life and social activities in Spain, such as drinking beer with friends at bars or enjoying wine with meals.  & Cultural Integration\\ 
   & [2] The legal drinking age in Spain is 18, with strict drink-driving laws (blood alcohol limit: 0.5 g/L).  & Legal and Social Norms\\  
   & [3] Spaniards practice moderate drinking, with binge drinking being uncommon. & Drinking Habits and Moderation\\
   & [4] Beer and wine are the most popular alcoholic beverages, with craft beer gaining popularity.  & Regional and Traditional Beverages\\
   \hline

    \multirow{5}{*}{Physical Contact} & [1] Spaniards emphasize physical contact as a tactile and emotional expression to convey friendliness, intimacy, or support.   & Frequency and Context of Physical Contact\\ 
   & [2] Public displays of affection (e.g., holding hands) between couples are widely accepted as normal in Spanish culture. & Cultural Acceptance and Boundaries\\  
   & [3] Friends use gestures like nudging arms, linking arms, or holding shoulders to show appreciation and support.  & Forms of Physical Contact\\
   & [4] Casual touches (e.g., adjusting clothes or lightly touching collars) signal approachability and friendliness.  & Forms of Physical Contact\\
   \hline

   \multirow{5}{*}{Widely Spoken Languages} & [1] Spanish is the official language in 20 countries and one U.S. territory (Puerto Rico).    & Official Status and Recognition\\ 
   & [2] Spanish is widely spoken in regions like the United States, parts of Africa, Asia, and Oceania.  ",
            & Geographical Distribution\\  
   &[3] The majority of Spanish speakers reside in Hispanic America, particularly Mexico (largest population of native speakers). & Geographical Distribution\\
   &[4] Spanish is recognized as an official/co-official language in international organizations (UN, EU, OAS, African Union). & Official Status and Recognition\\
   \hline
   
  \end{tabularx}
  \caption{Examples of extracted knowledge instances. Dimension refers to the finest-grained level used for retrieval and rewriting; these dimensions serve as keywords for Google search. Cluster category denotes the result of clustering the cultural knowledge, with each piece of knowledge corresponding to a more fine-grained category. }
  \label{tab:database}
\end{table*}

\section{Data Instances}
\label{sec: datainstance}
This section provides representative samples from the constructed dataset, covering both cultural knowledge instances and question examples in evaluation dataset. 

\subsection{Cultural Knowledge}
The extracted cultural knowledge instances are organized according to the evaluation dimensions. Each cultural knowledge instance provides a characteristic description of the corresponding cultural dimension as manifested in a specific cultural context, encompassing relevant commonsense knowledge, shared beliefs, values, and more. During dataset construction, we used the aforementioned fine-grained evaluation dimensions as cues for retrieval and rewriting. Cultural knowledge associated with the same dimension is grouped together. Furthermore, we clustered the knowledge instances under each dimension to derive even more fine-grained sub-dimensions, which can serve as references for future dataset construction and evaluation. 
We provide a detailed illustration of examples from the constructed knowledge bases in Table~\ref{tab:database}.

\subsection{Evaluation Dataset}
As is mentioned in the main context, the questions in our evaluation dataset are categorized into four types based on their content: Factual, Conceptual, Misleading, and Multi-hop reasoning. Each instance in the evaluation set consists of a question, its corresponding answer, and the associated cultural knowledge. The questions are categorized into subjective types (short answer and essay questions) and objective types (single-choice, multiple-choice, and true/false questions). Here, we present the examples organized by their content types in Table~\ref{tab:questions}.

\section{Prompts}
During the dataset construction process, we leveraged large language models (LLMs) for text rewriting, question generation, and quality assessment. In these steps, we employed carefully expert-designed prompts to guide the LLMs in performing the corresponding tasks. These prompts are language-specific and correspond to the language of the target data. Here, we present the prompts used across different languages.


\begin{table*}[t]
\centering
\renewcommand{\arraystretch}{1.2}
\begin{tabularx}{\textwidth}{|p{0.12\textwidth}|p{0.18\textwidth}|p{0.18\textwidth}|X|}
\hline
\textbf{Layer} & \textbf{Category} & \textbf{Topic Aspect}  & \textbf{Fine-grained Dimension} \\
\hline

\multirow{13}{*}{} & \multirow{9}{*}{} & \multirow{4}{*}{} & population rank \\
& & & population distribution \\
& & & land area percentage \\
& & & main regions \\
& & Population and & ethnicity \\
& & Geography & official languages \\
& Geography \& Customs & & widely spoken languages \\
& & & famous river \\
& & & climate \\
\cline{3-4}

& & & national holidays \\
& & & religious holidays \\
& & Dates of Significance & cultural holidays \\
& & & origin of festivals \\
& & & celebration of festivals \\
& & & symbol of festivals \\
\cline{2-4}

\multirow{20}{*}{\shortstack{Institutional\\Norms}} & \multirow{5}{*}{} &  & vehicle movement rules \\
& & & traffic signs and signals \\
& &Transportation Rules & pedestrian and non-motorized vehicles \\
& & & motor vehicle driving rules \\
& & & parking regulations \\
\cline{3-4}

& & & data order \\
& & & separator type \\
& & & year format \\
& &Data Format  & month representation \\
& & & zero-padding \\
& & & natural language date \\
& Regulation \& Policy & & calendar system \\
\cline{3-4}

& & Measurement Unit & measurement system \\
& & & localization \\
\cline{3-4}

& & & regulation structure \\
& & & market type \\
& & & entity type \\
& & & access \& licensing \\
& & & conduct \& compliance \\
& & Financial Market  & capital \& risk \\
& & Rules & monetary \& payment system \\
& & & finTech regulation \\
& & & tax \& accounting \\
& & & international alignment \\
& & & stability \& resolution \\
\cline{1-4}

\multirow{7}{*}{} & \multirow{7}{*}{} & & payment habits \\
& & & travel habits \\
& & Daily Life and Travel & bathing habits \\
& & & pet-raising habits \\
& & & social media usage pattern \\
& & & marriage customs \\
\cline{3-4}
& & Diet and Health & eating habits \\
& & Concept & attitudes towards birth, aging, illness and death \\
& & & traditional and modern medicine \\
\cline{3-4}
& & & educational practices \\
& & Education and  & teacher student concept \\
& & Knowledge & major selection \\
& & & career choices \\
\hline
\end{tabularx}
\end{table*}

\begin{table*}[t]
\centering
\renewcommand{\arraystretch}{1.2}
\begin{tabularx}{\textwidth}{|p{0.12\textwidth}|p{0.18\textwidth}|p{0.18\textwidth}|X|}
\hline
\textbf{Layer} & \textbf{Category} & \textbf{Topic Aspect}  & \textbf{Fine-grained Dimension} \\
\hline


& & & festive entertainment activities in different places with the same origin \\
& & Art and Entertainment & traditional music and dance \\
& & & contemporary art \\
\cline{3-4}

& & & basic etiquette \\
& & & naming convention \\
& & & name origin \\
& & & name significance \\
& & & punctuality when visiting \\
& & & shoe etiquette during a visit \\
& & & hospitality customs when receiving guests \\
& & & bringing gifts for the host \\
& & & seating etiquette for guests and hosts \\
& & & serving etiquette during a visit \\
& & & leaving food after eating \\
& & & using salt while eating \\
& & & giving compliments during a meal \\
& & & eating with the right hand \\
& & & alcohol \\
& Personal Choices  & & pork \\
& \& Habits  & & handing gifts \\
& & & gifts for children \\
& & & opening gifts \\
& & & general greeting principles \\
& & & greeting between men and men \\
Behavioral & & Etiquette and Courtesy & greeting between women and women \\
Patterns & & & greeting between men to women \\
& & & appointment scheduling before the business meeting \\
& & & dress code before the business meeting \\
& & & business card exchange before the business meeting \\
& & & network building before the business meeting \\
& & & age and experience before the business meeting \\
& & & familiarity before the business meeting \\
& & & socialization during the business meeting \\
& & & meeting duration during the business meeting \\
& & & open door policy during the business meeting \\
& & & interruptions during the business meeting \\
& & & deference to senior during the business meeting \\
& & & negotiation style during the business meeting \\
& & & decision making during the business meeting \\
& & & bartering during the business meeting \\
& & & private meetings during the business meeting \\
& & & confrontation avoidance during the business meeting \\
& & & follow up after the business meeting \\
& & & ongoing negotiations after the business meeting \\
& & & communication style \\
\hline

\end{tabularx}
\end{table*}

\begin{table*}[t]
\centering
\renewcommand{\arraystretch}{1.2}
\begin{tabularx}{\textwidth}{|p{0.12\textwidth}|p{0.18\textwidth}|p{0.18\textwidth}|X|}
\hline
\textbf{Layer} & \textbf{Category} & \textbf{Topic Aspect}  & \textbf{Fine-grained Dimension} \\
\hline

& & & indirect communication \\
& & & humour \\
& & & physical contact \\
& & & personal space \\
& & & gestures \\
& & & beckoning \\
& & & eye contact \\
\hline

& & & idiom \\
& & Fixed Expressions & common saying \\
& & in Language & proverb \\
& & & neologism and abbreviation \\
\cline{3-4}

& & Family Dynamics & communal living \\
& & & parental care \\
\cline{3-4}
& Social Relationship &  & patriarchy \\
& and Structures & & women roles in the family \\

& & Household Structures & social interaction \\
& & & urban rural gap \\
& & & concept of seniority and childhood \\
\cline{3-4}

& & & male dominance \\
& & Gender Roles & social compliance \\
Core Values& & & honour gender \\
and Social& & & changing attitudes of gender \\
\cline{2-4}

Structures & & & power distance \\
& & & collectivism \\
& & & individualism \\
& & Culture Values & motivation towards achievement and success \\
& & & uncertainty avoidance \\
& & & long term orientation \\
& & & indulgence \\
\cline{3-4}

& Values and & Religion & religion \\
\cline{3-4}

& Beliefs & & modest dress \\
& & & informality \\
& & & compliments \\
& & & cultural acknowledgement \\
& & & education \\
& & Do's and Don'ts & insults \\
& & & dirty jokes \\
& & & political criticism \\
& & & sensitive topics \\
& & & ethnicity assumptions \\
& & & stereotyping \\
\cline{3-4}

\hline
\end{tabularx}
\caption{The complete dimensional schema of CultureScope}
\label{tab:dims}
\end{table*}

\begin{table*}[t]

\centering
\renewcommand{\arraystretch}{1.15}
\begin{tabularx}{\textwidth}{|p{0.08\textwidth}|p{0.34\textwidth}|p{0.20\textwidth}|X|}
\hline
\textbf{Type} & \textbf{Question} & \textbf{Answer} & \textbf{Associated Knowledge} \\
\hline

Factual & 
During the festival of Las Fallas in Valencia, which element is prominently featured and plays a significant role in the celebrations?  
A) Ice sculptures  
B) Fireworks and bonfires  
C) Water fountains  
D) Sandcastles 
& 
B) Fireworks and bonfires 
& 
Use of Symbolic Elements: Symbolic elements such as fire, water, and specific foods play crucial roles in many Spanish festivals. Fire is especially prominent in celebrations like Las Fallas and San Juan’s Night. \\
\hline

Mislead & 
You are a foreign exchange student in Spain... *(long prompt truncated here for brevity in description — see full LaTeX for exact content)* ...Which of the following statements best reflects a critical understanding of Spanish culture?  
A) My friend's comment is accurate...  
B) My friend's comment reflects a common stereotype...  
C) My friend's comment is partially correct...  
D) My friend's comment is entirely incorrect...
& 
B) My friend's comment reflects a common stereotype, but it does not accurately represent the reality of modern Spanish families, where roles are more flexible and shared.
& 
Increasing numbers of Spanish women now pursue careers while balancing family duties.  
In Hispanic households, daughters are expected to help with chores, while sons participate less.  
In Spanish culture, women traditionally take primary responsibility for household chores and child-rearing. \\
\hline

Multi-hop & 
Imagine you are a Spanish student... *(long prompt truncated here)*  
1. Contextual Understanding: What could be the possible month...  
2. Cultural Practices: Why do you think shared meals...  
3. Language and Grammar: How would you abbreviate...
& 
1. Likely July (Julio) or August (Agosto), named after Julius Caesar and Augustus.  
2. Shared meals foster community, bonding, and family tradition.  
3. Abbreviations: "jul." / "ago."; lowercase in most grammatical uses. 
& 
Cultural significance of months: Julio (Julius Caesar), Agosto (Augustus); Sept–Dec = Latin numbers.  
Food as bonding ritual in Spanish culture.  
Grammar: months in lowercase unless at sentence start; masculine nouns; accepted abbreviations like “en.” for enero. \\
\hline

Conceptual & 
In a Spanish business context, you notice that after a formal meeting, colleagues often suggest coffee or drinks. What is the primary purpose?  
A) Finalize decisions  
B) Build personal trust  
C) Discuss new opportunities  
D) Reduce stress
& 
B) To build personal relationships and trust outside of the formal setting
& 
Informal gatherings (after-work events, coffee breaks) build trust and relationships outside formal settings. Personal bonds are often foundational before formal business discussions in Spain. \\
\hline
\end{tabularx}
\caption{Examples of generated evaluation data set. Based on content classification, the evaluation dataset comprises four distinct question types. Each question includes a prompt, a corresponding answer, and relevant cultural knowledge.}
\label{tab:questions}
\end{table*}

\FloatBarrier

\begin{tcolorbox}[colback=gray!5!white, colframe=gray!80!black, title=Knowledge Summary(English)]
I will provide a web article. Please extract the key characteristics and content related to the cultural dimension in Chinese culture from it. Present these features clearly under distinct headings.

Each feature should begin with a section titled:

Title
Then list the following points below:

Description of the feature:

Source of information: (Quote the original text and indicate the paragraph if possible.)

The content should be well-structured and logically coherent.
If the information is insufficient to support a certain feature, do not fabricate content.

The article is as follows:

===

\textit{Input Text}

===

\end{tcolorbox}

\begin{tcolorbox}[colback=gray!5!white, colframe=gray!80!black, title=Knowledge Summary(Spanish)]
Eres un investigador especializado en la cultura española. Se te proporcionará un texto de una página web (puede estar en cualquier idioma). 

Tu tarea es identificar las características culturales que estén relacionadas con el siguiente aspecto específico de la cultura española:

Dimensión cultural:**{dimension}**

Por favor, extrae del texto solo los elementos relevantes que estén claramente relacionados con esta dimensión cultural.

Escribe las características en español, siguiendo este formato:

[Título breve de la característica]  

[Descripción clara y concisa en español]

Fuente de información: [Frase, palabra clave o idea tomada directamente del texto original] 

Escribe siempre en español. .

Texto de la página web:

"""

\textit{Input text}

"""
\end{tcolorbox}









\begin{tcolorbox}[colback=gray!5!white, colframe=gray!80!black, title=Question Instruction(Factual)]
Based on the context, think through all relevant cultural points step by step and generate a \textbf{factual} question. The question type can include single-choice, true/false, or fill-in-the-blank. Ensure that the question stem is clear, the options are plausible but misleading (distractors), and the answer is accurate.
\end{tcolorbox}

\begin{tcolorbox}[colback=gray!5!white, colframe=gray!80!black, title=Question Instruction(Conceptual)]
Based on the context, think through all relevant cultural points step by step and generate a \textbf{conceptual} explanation question. The question should focus on the learner's understanding of the concepts, structures, or values behind cultural phenomena, rather than simple memorization. Suitable formats include multiple-choice or true/false questions. Ensure the question is thought-provoking and the answer is well-justified.

\end{tcolorbox}

\begin{tcolorbox}[colback=gray!5!white, colframe=gray!80!black, title=Question Instruction(Mislead)]
Based on the context, think through all relevant cultural points step by step and generate a \textbf{misleading} question to assess whether learners can identify cultural misunderstandings, stereotypes, or biases. The question should focus on learners' critical thinking about culture, identifying which statements or behaviors reflect misunderstandings, oversimplifications, biases, or stereotypes, and guide them toward more accurate or respectful understandings. Possible formats include multiple-choice, true/false, case analysis, or short-answer questions.

\end{tcolorbox}

\begin{tcolorbox}[colback=gray!5!white, colframe=gray!80!black, title=Question Instruction(Multi-hop)]
Based on the context, think through all relevant cultural points step by step and generate a \textbf{multi-hop} reasoning question to assess whether the learner can synthesize multiple cultural elements and understand the deeper logic or internal connections among cultural phenomena. The question should prompt learners to start from multiple information points, integrate cultural knowledge, and perform logical analysis, comparison, or generalization. Scenario-based, integrated analysis, or comparative reasoning questions are recommended.
\end{tcolorbox}





\begin{tcolorbox}[colback=gray!5!white, colframe=gray!80!black, title=Question Generation(English)]
Task: Answer in English. 
\textit{instruction}

Note:

1. The question should avoid explicitly mentioning cultural concepts, terminology, or characteristics, in order to effectively assess the student's understanding of cultural traits.

2. A reference answer should be provided after the question.

Context: 

'''

\textit{Context}

'''

Question:

Reference Answer:
\end{tcolorbox}



\begin{tcolorbox}[colback=gray!5!white, colframe=gray!80!black, title=Knowledge Injection(English)]

\textbf{Reference:}

\textit{reference knowledge}

\textbf{Question: }

\textit{question}
\end{tcolorbox}

\end{document}